%% file: iclr2027_conference.tex
\documentclass{article} 
\usepackage{iclr2027_conference,times}

\input{math_commands.tex}

\usepackage{hyperref}
\usepackage{url}

\usepackage{graphicx}
\usepackage[caption=false]{subfig}
\usepackage{booktabs}
\usepackage{multirow}

\title{EviLink: Multi-Path Schema Linking with Uncertainty-Guided Evidence Acquisition for Large-Scale Text-to-SQL}

\author{%
\textbf{Huawei Zheng\textsuperscript{1}\quad
Sen Yang\textsuperscript{1}\quad
Zhaorui Yang\textsuperscript{2}\quad
Yuhui Zhang\textsuperscript{3}\quad
Haozhe Feng\textsuperscript{3}}\\
\textbf{Haoxuan Li\textsuperscript{2}\quad
Xuan Yi\textsuperscript{4}\quad
Chao Hu\textsuperscript{3}\quad
Defeng Xie\textsuperscript{3}\quad
Chen Hou\textsuperscript{3}}\\
\textbf{Danqing Huang\textsuperscript{3}\quad
Wei Chen\textsuperscript{2}\quad
Peng Chen\textsuperscript{3\textdagger}\quad
Dazhen Deng\textsuperscript{1,2\textdagger}}\\[0.5em]
\normalfont\small
\textsuperscript{1}School of Software Technology, Zhejiang University\\
\normalfont\small
\textsuperscript{2}State Key Lab of CAD\&CG, Zhejiang University\\
\normalfont\small
\textsuperscript{3}Tencent TEG\\
\normalfont\small
\textsuperscript{4}School of Mathematical Sciences, Peking University\\[0.3em]
\normalfont\footnotesize
\textbf{Correspondence: }
\texttt{zhenghuawei@zju.edu.cn, aidenhzfeng@tencent.com}\\
\normalfont\footnotesize
\texttt{dengdazhen@zju.edu.cn}
}

\iclrfinalcopy 
\begin{document}

\maketitle

\fancyhead{}                       
\renewcommand{\headrulewidth}{0pt} 

\begingroup
\renewcommand{\thefootnote}{\fnsymbol{footnote}}
\setcounter{footnote}{2}
\footnotetext{%
\parbox[t]{0.92\linewidth}{%
Corresponding authors: Dazhen Deng and Peng Chen.\\
Under review. Code will be released upon acceptance.\\
This work was done during Huawei Zheng's internship at Tencent TEG Data Computing Platform Department.%
}}
\endgroup
\setcounter{footnote}{0}

\input{sections/0_abstract}
\input{sections/1_introduction}
\input{sections/2_related_work}
\input{sections/3_method}
\input{sections/4_experiments}
\input{sections/5_conclusion}

\bibliography{iclr2027_conference}
\bibliographystyle{iclr2027_conference}

\appendix

\input{appendix/baseline}
\input{appendix/gold}
\input{appendix/metric}
\input{appendix/settings}

\input{appendix/table_results}
\input{appendix/size}
\input{appendix/more_models}

\input{appendix/multiple}
\input{appendix/sensitive}
\input{appendix/refinement}

\input{appendix/tool}
\input{appendix/prompt}

\end{document}

%% file: math_commands.tex
\usepackage{amsmath,amsfonts,bm}

\def\eqref#1{equation~\ref{#1}}

\def\1{\bm{1}}

\DeclareMathAlphabet{\mathsfit}{\encodingdefault}{\sfdefault}{m}{sl}
\SetMathAlphabet{\mathsfit}{bold}{\encodingdefault}{\sfdefault}{bx}{n}



%% file: sections/0_abstract.tex
\begin{abstract}
Schema linking is a difficult and important step in large-scale Text-to-SQL, where systems must identify a compact yet sufficient schema context from large and ambiguous databases. Existing methods often treat schema linking as deterministic selection around a single SQL path, but complex questions may admit multiple valid realizations with different schema needs. We reframe schema linking as uncertainty-aware schema-need inference over multiple plausible SQL paths, where the system distinguishes required schema items from path-dependent uncertain ones and acquires evidence only where needed. We instantiate this reframing with EviLink, which combines multi-hypothesis schema grounding with uncertainty-guided evidence acquisition. Experiments on BIRD-Dev and Spider2-Snow show that this perspective improves the balance among schema completeness, schema relevance, and token cost. On Spider2-Snow, EviLink achieves 93.04\% field-level strict recall rate, uses 116.55K average tokens, and improves downstream SQL generation under a fixed generator.
\end{abstract}

%% file: sections/1_introduction.tex
\section{Introduction}

Text-to-SQL translates natural language questions into executable SQL queries, making database analysis accessible to non-expert users \citep{yu-etal-2018-spider,nl2sql360}. Despite recent progress driven by large language models \citep{NEURIPS2023_72223cc6,10.14778/3641204.3641221}, complex database scenarios still face a fundamental bottleneck. Before generating SQL, the system must identify a compact yet sufficient schema context \citep{wang-etal-2025-linkalign}. This step, known as \emph{schema linking}, determines which tables and fields are available to the SQL generator \citep{wang-etal-2020-rat,maamari2024deathschemalinkingtexttosql}. High-quality schema linking reduces irrelevant schema noise, alleviates context-window pressure, and provides the information basis for faithful SQL generation \citep{cao2024rslsqlrobustschemalinking}. Recent work further suggests that strong schema linking can help a relatively simple agentic backend achieve promising end-to-end Text-to-SQL performance \citep{Wang_Zheng_Cao_Zhang_Wei_Fu_Luo_Chen_Bai_2026}.

\begin{figure}[t]
    \centering
    \includegraphics[width=\linewidth]{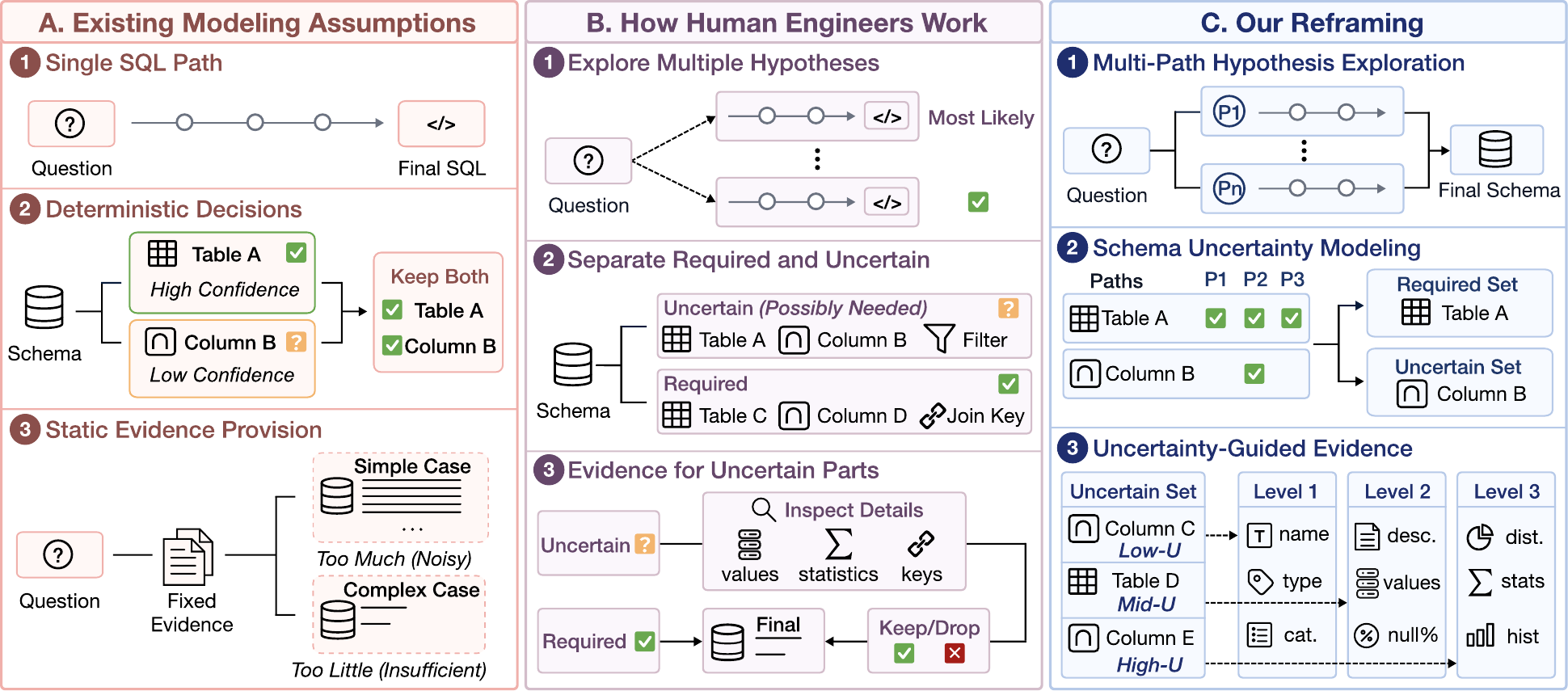}
    \caption{\textbf{From single-path schema linking to uncertainty-guided multi-path reasoning.} (A) Many existing formulations rely on a single SQL path, deterministic schema decisions, and static evidence provision. (B) Human engineers instead consider multiple plausible SQL paths, separate required from uncertain schema items, and inspect evidence selectively. (C) We reframe schema linking as uncertainty-guided multi-path reasoning, which preserves multiple solution paths, models schema uncertainty, and uses targeted evidence to resolve uncertain schema needs.}
    \label{fig:motivation}
\end{figure}

Existing methods usually formulate schema linking as selecting or ranking schema items relevant to a question \citep{Li_Zhang_Li_Chen_2023,cao2024rslsqlrobustschemalinking}. Earlier methods model schema-question alignment with semantic matching, structured encoders, or learned rankers \citep{wang-etal-2020-rat,Li_Zhang_Li_Chen_2023}. Recent LLM-based and agentic methods further improve schema selection through SQL-draft-based pruning, contextual augmentation, database retrieval, or schema exploration \citep{cao2024rslsqlrobustschemalinking,wang-etal-2025-linkalign,Wang_Zheng_Cao_Zhang_Wei_Fu_Luo_Chen_Bai_2026,cao2026apexsqltalkingdataagentic}. However, as illustrated in Figure~\ref{fig:motivation}, many existing formulations still reduce schema linking to selecting a final schema subset under assumptions that can become limiting in complex settings. They tend to reason around a single SQL path, treat schema items as deterministic keep-or-drop decisions, or allocate schema evidence in a largely fixed way.

We argue that these assumptions limit schema linking in complex Text-to-SQL, especially in realistic enterprise-scale benchmarks \citep{ICLR2025_46c10f6c}. A question may have multiple valid SQL implementations \citep{bhaskar-etal-2023-benchmarking,NEURIPS2024_a4c942a8}. These implementations can use different join paths, bridge tables, filtering fields, aggregation choices, or intermediate query structures. Since different SQL paths may require different schema elements, linking schema around a single assumed path can prematurely discard tables or fields needed by other valid solutions. Once removed, these elements become unavailable to the downstream SQL generator, causing irreversible information loss. Schema linking should therefore infer schema needs across plausible SQL paths, rather than merely recover the schema appearing in one reference SQL.

This multi-path view also changes how schema uncertainty and evidence should be handled. Some schema elements are consistently required, some are clearly irrelevant, and others are conditionally needed depending on the SQL path. However, existing formulations often collapse this uncertainty into deterministic keep/drop decisions or fixed ranking thresholds. Static evidence makes this problem worse. Lightweight evidence may be insufficient for difficult cases, while detailed evidence for all schema elements adds redundant context and token cost. For interactive data agents, token efficiency is therefore a practical concern alongside schema completeness. As Figure~\ref{fig:motivation} suggests, a more faithful schema linker should distinguish required and uncertain schema items, and acquire additional evidence only where uncertainty remains.

Motivated by this observation, we introduce \textbf{EviLink} to instantiate this uncertainty-aware view of schema linking. EviLink first constructs multiple plausible SQL hypotheses. It then estimates schema support across these hypotheses to separate required schema items from uncertain ones, and allocates stronger evidence only to uncertain items. By jointly modeling path diversity, schema uncertainty, and targeted evidence acquisition, EviLink preserves schema elements needed by alternative SQL paths while avoiding the noise and cost of uniformly exposing detailed evidence.

Our contributions are summarized as follows:
\begin{itemize}
    \item \textbf{Problem reframing.} We reframe schema linking for complex Text-to-SQL as schema-need inference over multiple plausible SQL hypotheses, rather than deterministic selection around a single assumed SQL path. This reframing highlights three limiting assumptions in existing formulations: single-path reasoning, deterministic schema decisions, and static schema evidence.

    \item \textbf{Method instantiation.} Concretely, we instantiate this reframing as EviLink, which combines multi-hypothesis schema grounding, required/uncertain schema bucketing, and uncertainty-guided evidence acquisition to preserve alternative schema needs while controlling irrelevant context.

    \item \textbf{Systematic validation.} We evaluate EviLink on BIRD-Dev and Spider2-Snow, achieving high schema completeness and reducing schema-linking tokens by 49.6\% to 79.9\% relative to APEX-SQL, with slightly higher execution accuracy.
\end{itemize}

%% file: sections/2_related_work.tex
\section{Related Work}

Prior schema-linking methods formulate the task as selecting or ranking schema items relevant to a question. Early neural methods model question--schema interactions with structured encoders, semantic matching, rankers, or contextual schema representations \citep{wang-etal-2020-rat,cao-etal-2021-lgesql,Li_Zhang_Li_Chen_2023}. Recent LLM-based Text-to-SQL methods improve schema selection and schema-aware generation through prompting, in-context demonstrations, schema augmentation, or SQL-draft-based refinement \citep{NEURIPS2023_72223cc6,dong2023c3zeroshottexttosqlchatgpt,10.14778/3641204.3641221,lee-etal-2025-mcs,qu-etal-2024-generation,liu-etal-2025-solid,yang2024sqltoschemaenhancesschemalinking,cao2024rslsqlrobustschemalinking,10.1145/3654930,talaei2024chesscontextualharnessingefficient}. More recent systems introduce interaction and exploration into schema linking. LinkAlign~\citep{wang-etal-2025-linkalign} separates database retrieval from schema item grounding, AutoLink~\citep{Wang_Zheng_Cao_Zhang_Wei_Fu_Luo_Chen_Bai_2026} performs iterative retrieval, exploration, verification, and expansion, ReFoRCE~\citep{deng2025reforcetexttosqlagentselfrefinement} and DSR-SQL~\citep{hao2025texttosqldualstatereasoningintegrating} incorporate field exploration or adaptive context, and APEX-SQL~\citep{cao2026apexsqltalkingdataagentic} uses logical planning and data-grounded verification.

These methods advance schema linking, but they generally use interaction, exploration, or evidence to improve one consolidated schema context. In contrast, EviLink treats schema needs as SQL-path dependent, since different plausible SQL realizations may use different join paths, filters, aggregations, or intermediate structures \citep{bhaskar-etal-2023-benchmarking,NEURIPS2024_a4c942a8}. It keeps path-wise schema needs explicit before consolidation, estimates cross-path support, separates required from uncertain items, and acquires additional evidence only where uncertainty remains. Overall, EviLink reframes schema linking as uncertainty-aware schema-need inference rather than single-context schema selection.

%% file: sections/3_method.tex
\section{Methodology}
\label{sec:method}

\begin{figure*}[t]
  \centering
  \includegraphics[width=\textwidth]{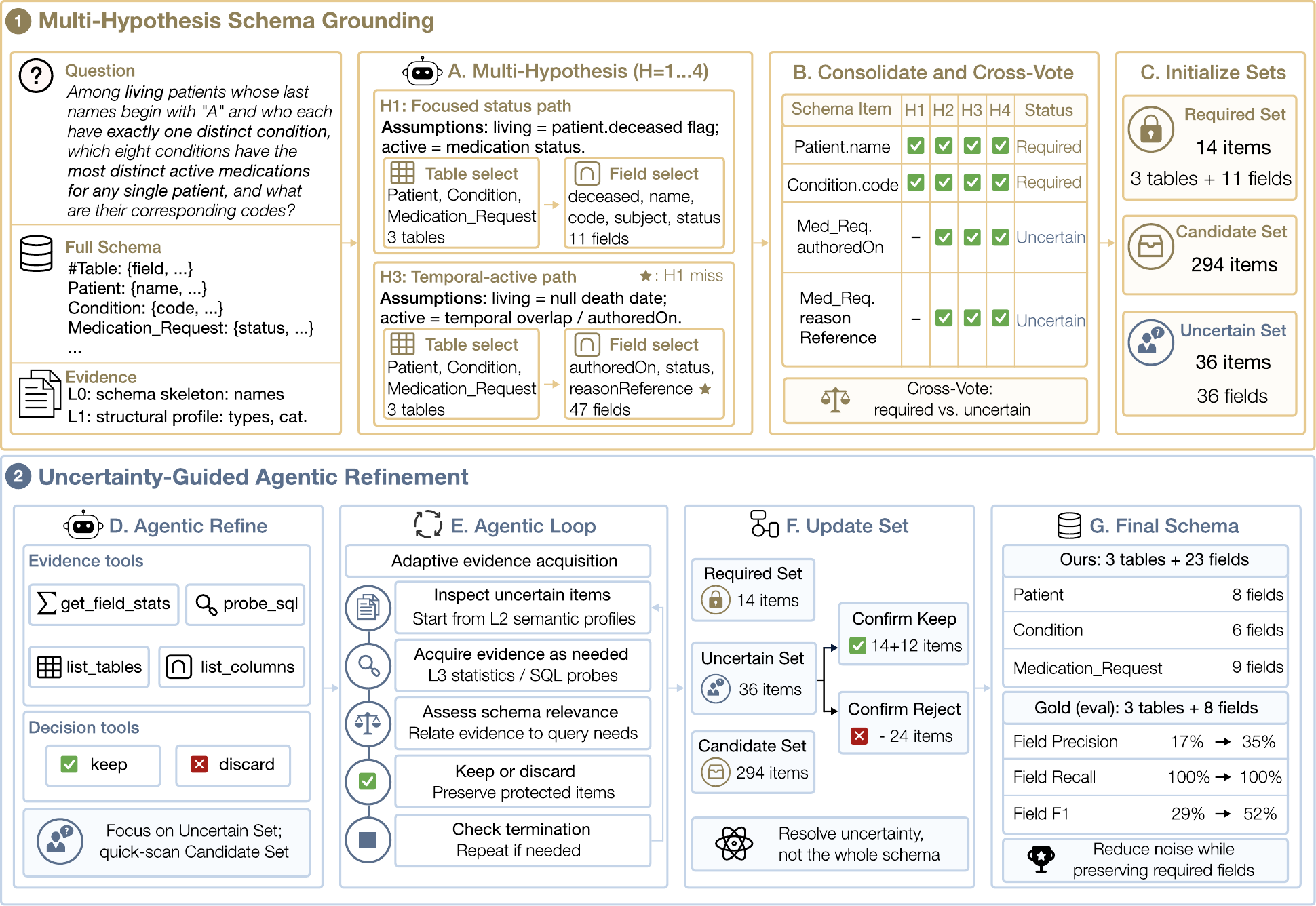}
  \caption{\textbf{Overview of EviLink.} The first stage performs multi-hypothesis schema grounding: (A) Multi-Hypothesis proposes multiple plausible SQL paths, (B) Consolidate and Cross-Vote aggregates their schema selections, and (C) Initialize Sets separates required, uncertain, and candidate items. The second stage performs uncertainty-guided agentic refinement: (D) Agentic Refine provides evidence and decision tools, (E) Agentic Loop resolves uncertain items with targeted evidence, (F) Update Set applies keep/discard decisions, and (G) Final Schema returns a compact schema while preserving required fields.}
  \label{fig:pipeline}
\end{figure*}

\paragraph{Task definition.}
Given a natural language question $q$ and a database schema $\mathcal{S}$, schema linking aims to identify a subset of schema items $\hat{\mathcal{S}} \subseteq \mathcal{S}$ as the schema context for downstream SQL generation. In this work, schema items include both tables and fields. The objective is to preserve items that may be required for producing a valid SQL query, while filtering out clearly irrelevant ones. Unlike SQL generation, schema linking does not produce an executable query; instead, it determines which parts of the schema are made available to the SQL generator.

\paragraph{Overview.}
Figure~\ref{fig:pipeline} shows the workflow of EviLink. The method proceeds in two stages. First, \emph{multi-hypothesis schema grounding} (Figure~\ref{fig:pipeline}, A--C) explores multiple SQL paths for the question instead of committing to a single path early. Each path selects tables and fields under lightweight evidence, and the path-wise selections are consolidated into required, uncertain, and candidate sets. Second, \emph{uncertainty-guided agentic refinement} (UGAR; Figure~\ref{fig:pipeline}, D--G) focuses the agent on the uncertain set rather than the whole database. The agent acquires stronger profiling or execution evidence as needed, keeps supported items, discards unsupported distractors, and returns a compact schema context for downstream SQL generation.

\subsection{Multi-Hypothesis Schema Grounding}

The first phase constructs a recall-oriented schema subset for later refinement. Rather than making final keep/drop decisions upfront, EviLink collects tables and fields that may be required by different plausible SQL realizations, reducing the risk of prematurely pruning alternative valid paths.

\paragraph{Multi-hypothesis path construction.}
We ask the model to generate a set of structured hypotheses $\mathcal{H}=\{h_1,\ldots,h_M\}$, where $M \leq K$ is adaptively chosen and $K$ is the maximum hypothesis budget. Each hypothesis describes one plausible SQL realization of the question, including relevant entities, filters, joins, aggregations, and ambiguous schema interpretations. The hypotheses should be structurally diverse, not superficial paraphrases. They should differ in at least one SQL-relevant dimension, such as join route, filter encoding, aggregation granularity, or temporal interpretation. For simple questions, the model may output a single hypothesis; for ambiguous questions, it may output multiple structurally distinct hypotheses.

\paragraph{Hypothesis-conditioned schema grounding.}
For each hypothesis $h_i$, we run an independent schema grounding path. The path first selects tables from the full catalog and then selects fields from the retained tables:
\begin{equation}
\label{eq:table-selection}
\mathcal{T}_i = \mathrm{SelectTable}(q,h_i,\mathcal{S}),
\end{equation}
\begin{equation}
\label{eq:field-selection}
\mathcal{F}_i = \mathrm{SelectField}(q,h_i,\mathcal{T}_i).
\end{equation}
Both selectors are recall-oriented: a table or field is retained if it may be required by any plausible SQL query under the current hypothesis. When uncertain, the selector is instructed to keep the item rather than discard it, which helps preserve bridge tables, join keys, temporal fields, and alternative encodings of the same semantic condition.

To keep grounding affordable on large schemas, table selection is performed over a compact schema view, and field selection is applied only to the retained tables. The hypothesis guides selection but does not act as a whitelist: schema items not explicitly mentioned in the hypothesis can still be selected when compact profiling signals indicate potential relevance.

\paragraph{Cross-voting and set initialization.}
After grounding all hypotheses, EviLink consolidates the path-wise outputs through cross-voting. Let $\mathcal{S}_i=\mathcal{T}_i\cup\mathcal{F}_i$ denote the schema items selected under hypothesis $h_i$. For each schema item $x \in \mathcal{S}$, we define its path support as
\begin{equation}
\label{eq:path-support}
n_x =
\left|
\left\{
i \in \{1,\ldots,M\}
\mid
x \in \mathcal{S}_i
\right\}
\right|.
\end{equation}
Items supported by most hypotheses are likely to be required, while items supported by only part of the hypotheses reflect path-dependent uncertainty.

Because the number of generated hypotheses is small, we smooth this support signal with a uniform Beta prior. We model the latent selection tendency of item $x$ as $p_x$:
\begin{equation}
\label{eq:posterior}
p_x \mid n_x,M
\sim
\mathrm{Beta}(n_x+1,M-n_x+1),
\end{equation}
and define its credibility score as
\begin{equation}
\label{eq:credibility}
c_x = \Pr(p_x > 0.5 \mid n_x,M).
\end{equation}
We use $c_x$ as a smoothed measure of support within the generated hypothesis set, rather than a calibrated probability that $x$ is necessary for every valid SQL realization. EviLink then initializes three sets. For $M\geq2$, items with $c_x \geq \tau_{\mathrm{req}}$ are placed into the required set $\mathcal{S}_{\mathrm{req}}$ and locked during refinement, while items with $0<n_x$ but $c_x < \tau_{\mathrm{req}}$ are placed into the uncertain set $\mathcal{S}_{\mathrm{unc}}$. For $M=1$, all selected items are instead locked in $\mathcal{S}_{\mathrm{req}}$ to preserve the single grounding path. Items with zero support remain in the candidate set $\mathcal{S}_{\mathrm{cand}}$, which is not refined by default but can still be inspected for recall-oriented recovery. In our implementation, the maximum hypothesis budget is $K=4$ and $\tau_{\mathrm{req}}=0.85$; when four hypotheses are generated, an item supported by three hypotheses is still treated as uncertain rather than required. Appendix~\ref{app:k_sensitivity} analyzes sensitivity to the hypothesis budget $K$.

\subsection{Uncertainty-Guided Agentic Refinement}
\label{sec:agentic-refinement}

The grounding stage is intentionally conservative: it preserves schema items that may be needed by at least one plausible SQL path. Refinement therefore focuses on resolving path-dependent uncertainty rather than pruning the full database again.

\paragraph{Evidence hierarchy.}
EviLink allocates profiling evidence according to schema uncertainty. We organize evidence into four levels:
\begin{itemize}
    \item \textbf{Schema skeleton\textnormal{ (L0)}} contains table and field names;
    \item \textbf{Structural profile\textnormal{ (L1)}} augments L0 with field types and coarse semantic categories;
    \item \textbf{Semantic profile\textnormal{ (L2)}} provides descriptions, representative values, and compact value-level summaries;
    \item \textbf{Statistical profile\textnormal{ (L3)}} stores additional field statistics when available, such as ranges, summaries, histograms, and distributional signals.
\end{itemize}
The grounding component uses L0 for broad table selection and L1 for field selection. During refinement, uncertain items are shown with L2 evidence by default, and L3 evidence is acquired only through \texttt{get\_field\_stats} when compact evidence is insufficient. Incomplete or noisy profiles may affect refinement decisions; unresolved uncertain items are retained by default to limit unsupported pruning.

\paragraph{Tool-based refinement.}
The refinement agent uses two types of tools. Evidence tools are read-only: \texttt{list\_tables} and \texttt{list\_columns} retrieve schema context for uncertain items or candidate recovery, \texttt{get\_field\_stats} upgrades selected fields to L3 and returns available evidence, and \texttt{probe\_sql} tests whether a hypothesized schema usage is feasible. Decision tools update the refinement state, where \texttt{keep} confirms an uncertain or recovered candidate item and \texttt{discard} removes an uncertain item with an explicit reason.

Schema items can move between sets through three updates:
\begin{equation}
S_{\mathrm{unc}} \stackrel{\mathtt{keep}}{\longrightarrow} S_{\mathrm{keep}}, \qquad
S_{\mathrm{unc}} \stackrel{\mathtt{discard}}{\longrightarrow} S_{\mathrm{drop}}, \qquad
S_{\mathrm{cand}} \stackrel{\mathtt{keep}}{\longrightarrow} S_{\mathrm{keep}}.
\end{equation}
Here, $S_{\mathrm{keep}}$ and $S_{\mathrm{drop}}$ contain items confirmed as kept and discarded, respectively. Evidence acquisition remains read-only, while schema updates follow agent decisions and predefined protection and recovery rules.

\paragraph{Convergence and final schema.}
The uncertain-item loop terminates when the uncertain set becomes empty, when the agent stops requesting tools, or when its turn budget is reached; candidate recovery is handled separately. Let $\mathcal{S}_{\mathrm{keep}}$ and $\mathcal{S}_{\mathrm{drop}}$ denote the items confirmed as kept and discarded, respectively, and let $\mathcal{S}_{\mathrm{pending}}$ denote uncertain items that remain unresolved at termination. To preserve the recall-first property of the grounding stage, unresolved uncertain items are kept by default. The linked schema returned to the downstream SQL generator is
\begin{equation}
\label{eq:final-schema}
\hat{\mathcal{S}}
=
\mathcal{S}_{\mathrm{req}}
\cup
\mathcal{S}_{\mathrm{keep}}
\cup
\mathcal{S}_{\mathrm{pending}} .
\end{equation}
Items in $\mathcal{S}_{\mathrm{drop}}$ are excluded from the final schema. This fallback does not override the agent's decisions; it only specifies how unresolved uncertainty is handled under the turn budget. As a result, refinement improves precision by rejecting unsupported uncertain items while avoiding aggressive pruning of schema items that may still be required by a valid SQL path.

%% file: sections/4_experiments.tex
\section{Experiments}

\subsection{Experimental Setup}
\label{sec:experimental_setup}

\paragraph{Benchmarks.}
We evaluate our method on two widely used Text-to-SQL benchmarks, BIRD-Dev~\citep{NEURIPS2023_83fc8fab} and Spider2-Snow~\citep{ICLR2025_46c10f6c}. BIRD-Dev provides a broad testbed for database question answering, while Spider2-Snow targets enterprise-level workflows with greater challenges in schema scale, schema ambiguity, and realistic data access. Together, these benchmarks enable evaluation across different database scales and task complexities. We conduct detailed experiments on both benchmarks to assess schema completeness, schema relevance, and token cost, as well as the impact on downstream SQL generation.

\paragraph{Baselines.}
We compare against seven recent Text-to-SQL and schema-linking baselines: TA-SQL~\citep{qu-etal-2024-generation}, RSL-SQL~\citep{cao2024rslsqlrobustschemalinking}, ReFoRCE~\citep{deng2025reforcetexttosqlagentselfrefinement}, LinkAlign~\citep{wang-etal-2025-linkalign}, AutoLink~\citep{Wang_Zheng_Cao_Zhang_Wei_Fu_Luo_Chen_Bai_2026}, DSR-SQL~\citep{hao2025texttosqldualstatereasoningintegrating}, and APEX-SQL~\citep{cao2026apexsqltalkingdataagentic}. These baselines span methods centered on schema linking and broader Text-to-SQL systems that explicitly model schema usage through linking, pruning, or exploration modules. For a fair and reproducible comparison, we adapt the official implementations only for input formatting, a unified model interface, and output normalization for evaluation. We preserve their method-defining prompts, decision procedures, retrieval and selection logic, and thresholds. Further details on the baselines and comparison scope are provided in Appendix~\ref{app:benchmarks_baselines}.

\paragraph{Models.}
All methods use the same backbone within each benchmark for the main schema-linking comparison: Qwen3.6-35B-A3B on BIRD-Dev and GLM-5.1 on Spider2-Snow. Schema-linking ablations use Qwen3.8-27B on both benchmarks. Downstream SQL generation uses Qwen3.8-27B on BIRD-Dev and Qwen3.6-27B on Spider2-Snow, with generators fixed across methods within each benchmark and schema contexts taken from the corresponding main linking runs. Appendix~\ref{app:additional_backbone} reports an additional backbone comparison. Detailed parameter settings are provided in Appendix~\ref{app:parameter}.

\paragraph{Evaluation metrics.}
Following RSL-SQL~\citep{cao2024rslsqlrobustschemalinking} and APEX-SQL~\citep{cao2026apexsqltalkingdataagentic}, we evaluate schema linking as a retrieval problem using four complementary metrics: \textbf{Strict Recall Rate (SRR)}, \textbf{Non-Strict Recall (NSR)}, \textbf{Non-Strict Precision (NSP)}, and \textbf{Non-Strict F1-score (NSF)}. To ensure reliable evaluation, we derive gold tables and fields from reference SQL through abstract syntax tree parsing and multi-agent cross-checking on both benchmarks, with additional reference SQL construction, execution verification, and targeted human SQL review for Spider2-Snow; details are provided in Appendix~\ref{app:gold}. These metrics can be computed with either tables or fields as retrieval units, and the exact computation procedure is provided in Appendix~\ref{app:metric}. All four metrics are reported as percentages. In the main experiments, we report field-level results by default because field-level evaluation is more fine-grained and stricter; table-level results are included in Appendix~\ref{app:table_results} for reference. \textbf{SRR} measures strict schema completeness: an example is counted as correct only when all gold units at the chosen granularity are retained. Following prior work, we treat SRR as the primary recall-oriented metric, since missing any required field can make the correct SQL unreachable. \textbf{NSR} measures average schema coverage, i.e., the average fraction of gold units recovered per example. \textbf{NSP} measures average predicted-schema relevance, i.e., the average fraction of predicted units that are relevant per example. \textbf{NSF} measures the average per-example balance between coverage and relevance. Following prior work, we report \textbf{Avg. Tokens} to quantify each method's token cost. We sum API-reported input and output tokens over all LLM calls, separately for schema linking and SQL generation. Prompts, evidence, tool outputs, revisions, and execution feedback are counted whenever included in an LLM call. Other cost metrics, such as latency, tool-call counts, and monetary costs, are less directly comparable due to differences in hardware, tool design, and provider pricing.

\input{tables/mainstream}

\subsection{Main Results}
\label{sec:main_results}

Table~\ref{tab:field_schema_linking_results} reports the field-level schema linking results on BIRD-Dev and Spider2-Snow. The two benchmarks allow us to examine EviLink under different database scales and complexity levels.

On BIRD-Dev, EviLink achieves a field-level SRR of 93.22\%, surpassed only by LinkAlign and AutoLink, while maintaining substantially higher precision than both. LinkAlign and AutoLink achieve SRR scores of 95.11\% and 98.89\%, respectively, but their NSP scores are only 10.68\% and 11.39\%, indicating substantial over-retrieval. In comparison, EviLink achieves 51.58\% NSP and 63.60\% NSF while retaining 98.05\% NSR, providing high schema coverage with considerably less irrelevant context. EviLink also outperforms RSL-SQL and APEX-SQL across all four schema-linking metrics. Its average token consumption is 25.68K, compared with 51.81K for AutoLink and 50.95K for APEX-SQL. These results demonstrate a favorable balance among schema coverage, relevance, and token cost on BIRD-Dev.

On Spider2-Snow, where databases are larger and closer to enterprise-style settings, EviLink shows its clearest recall advantage. It achieves the best recall among all methods, with 93.04\% SRR and 97.86\% NSR, while using 116.55K tokens on average. Compared with APEX-SQL, EviLink improves SRR from 80.85\% to 93.04\% and NSR from 95.26\% to 97.86\%, while reducing token usage from 580.80K to 116.55K. However, its NSP and NSF are lower than those of APEX-SQL, at 26.04\% and 36.65\% versus 47.50\% and 58.64\%, respectively. These results show that EviLink's main advantage in this setting lies in schema completeness and linking efficiency. Appendix~\ref{app:schema_size} further shows that EviLink achieves the highest field-level SRR on medium and large schemas, exceeding the strongest baseline in each group by 5.06 and 14.29 percentage points, respectively. On the 95-example subset with unchanged officially released SQL, EviLink also achieves the highest field-level SRR (91.58\%) and NSR (98.17\%) with the lowest linking token consumption (55.27K). These results further support EviLink's recall and token-efficiency advantages (Appendix~\ref{app:official_sql_subset}).

\subsection{Ablation Study}
\label{sec:ablation}

Table~\ref{tab:ablation} reports ablation results on BIRD-Dev and Spider2-Snow using Qwen3.8-27B for all variants. We evaluate on all 1,534 BIRD-Dev examples and the 517 gold-labeled Spider2-Snow instances. On Spider2-Snow, removing multi-hypothesis grounding reduces SRR from 74.66\% to 63.25\% and NSR from 91.54\% to 86.78\%, while NSP and NSF increase from 54.58\% to 60.56\% and from 63.32\% to 66.77\%, respectively. On BIRD-Dev, removing this component similarly reduces SRR from 92.57\% to 79.92\% and NSR from 97.82\% to 93.61\%, while increasing NSP and NSF. This suggests that multi-hypothesis grounding mainly improves schema coverage by preserving alternative SQL paths, at the cost of lower precision and higher token consumption.

Uncertainty-guided refinement shows a complementary effect, providing modest but consistent improvements in schema coverage and relevance. Removing refinement decreases all four field-level metrics on both benchmarks. On Spider2-Snow, SRR drops from 74.66\% to 74.47\% and NSR from 91.54\% to 91.26\%, while on BIRD-Dev, NSP decreases from 56.13\% to 55.12\%. Relative to \textit{w/o UGAR}, Full improves NSF by 1.04 percentage points on BIRD-Dev and 0.85 on Spider2-Snow, with additional token consumption of 2.87K and 11.93K, respectively. Appendix~\ref{app:ugar_analysis} further shows that refinement's field F1 gains are more pronounced on Spider2-Snow examples whose schemas change, with cases illustrating both selective pruning and recovery.

Adaptive evidence acquisition provides a practical recall--cost trade-off. Static L0--L3 each use one evidence level throughout. Static L0 uses limited evidence and obtains lower SRR on both benchmarks, although its higher NSP and NSF reflect a more selective schema context. On Spider2-Snow, Static L1 and Static L2 achieve higher SRR than Full, but require 179.63K and 521.07K tokens, respectively, compared with 91.36K for Full. Static L2 reaches 78.34\% SRR, while its NSF of 63.03\% remains slightly below Full's 63.32\%. Static L3 consumes 1991.14K tokens, more than \(21\times\) Full, while its NSF is only 0.30 percentage points higher and its SRR is lower (74.08\% vs. 74.66\%). On BIRD-Dev, Full achieves 92.57\% SRR with 28.16K tokens, while Static L1 reaches 92.63\% with 37.18K tokens. Static L1--L3 obtain higher NSF at greater token cost. These results support uncertainty-guided evidence allocation as a cost-efficient way to retain high schema coverage, although it does not maximize every retrieval metric.

\input{tables/ablation}

\subsection{Downstream SQL Generation}
\label{sec:downstream_sql_generation}

To examine whether schema-linking quality translates into better end-to-end SQL generation, we follow the downstream evaluation protocol of AutoLink~\citep{Wang_Zheng_Cao_Zhang_Wei_Fu_Luo_Chen_Bai_2026} on BIRD-Dev ($N=1534$) and Spider2-Snow ($N=517$), using schema inputs from the corresponding main linking runs. We reuse its pipeline and prompts, sampling three SQL candidates, allowing up to three execution-guided revision rounds per candidate, and selecting the final SQL by execution-based voting. Within each benchmark, only the linked schema varies across methods; the question, database, prompt templates, and SQL generator remain fixed.

As shown in Table~\ref{tab:sql_generation_results}, EviLink achieves the highest execution accuracy (EX), measured by execution-result agreement with gold answers, on both benchmarks. With the SQL generator fixed within each benchmark, it obtains 66.23\% EX on BIRD-Dev and 39.07\% on Spider2-Snow, exceeding APEX-SQL by modest margins of 0.46 and 0.58 percentage points, respectively. Appendix~\ref{app:multiple} further illustrates how different evaluator-accepted SQL paths can induce different schema requirements.

\begin{figure*}[t]
  \centering
  \subfloat[BIRD-Dev\label{fig:ex_bird}]{
    \includegraphics[width=0.48\textwidth]{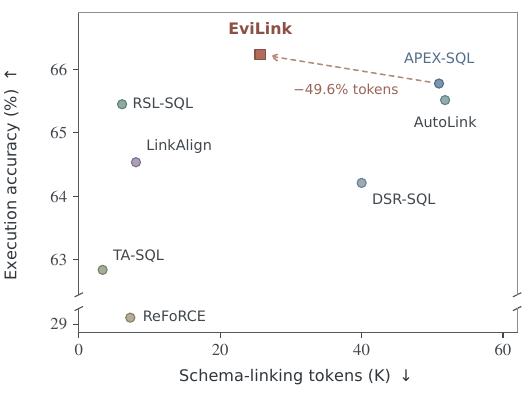}
  }
  \hfill
  \subfloat[Spider2-Snow\label{fig:ex_spider}]{
    \includegraphics[width=0.48\textwidth]{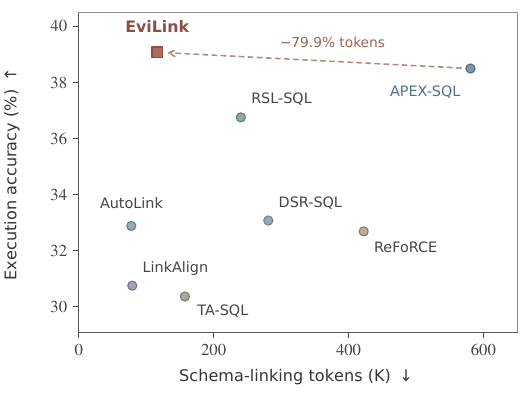}
  }
  \caption{Downstream SQL generation performance on BIRD-Dev and Spider2-Snow.}
  \label{fig:ex}
\end{figure*}

Figure~\ref{fig:ex} shows these accuracy gains alongside linking-token reductions of 49.6\% on BIRD-Dev and 79.9\% on Spider2-Snow relative to APEX-SQL. Generation tokens are slightly lower on BIRD-Dev (2.83K vs. 2.90K) but higher on Spider2-Snow (50.17K vs. 45.53K). Overall, total tokens decrease by approximately 47.1\% and 73.4\%, respectively.

\input{tables/sqlgen}

%% file: tables/mainstream.tex
\begin{table*}[t]
\caption{Field-level schema linking performance on BIRD-Dev and Spider2-Snow. Token denotes the average per-example token consumption during schema linking. Best results within each dataset are shown in bold. Results on BIRD-Dev are obtained with Qwen3.6-35B-A3B, while results on Spider2-Snow are obtained with GLM-5.1.}
\label{tab:field_schema_linking_results}
\centering

{
\setlength{\tabcolsep}{4.0pt}

\begin{tabular}{@{}lccccccccccc@{}}
\toprule
\multirow{2}{*}{\textbf{Method}}
& \multicolumn{5}{c}{\textbf{BIRD-Dev} ($N=1534$)}
&
& \multicolumn{5}{c}{\textbf{Spider2-Snow} ($N=517$)} \\
\cmidrule(lr){2-6} \cmidrule(lr){8-12}
& SRR & NSR & NSP & NSF & Token
&
& SRR & NSR & NSP & NSF & Token \\
\midrule

TA-SQL
& 73.73 & 89.84 & \textbf{81.91} & \textbf{83.98} & \textbf{3.38K}
& & 42.94 & 73.12 & \textbf{70.02} & \textbf{68.13} & 157.41K \\

RSL-SQL
& 89.90 & 97.10 & 44.32 & 57.89 & 6.13K
& & 79.50 & 95.22 & 24.71 & 34.48 & 240.34K \\

ReFoRCE
& 20.01 & 54.61 & 71.87 & 58.64 & 7.29K
& & 40.43 & 81.91 & 50.01 & 55.48 & 422.47K \\

LinkAlign
& 95.11 & 98.51 & 10.68 & 18.05 & 8.09K
& & 63.44 & 80.70 & 10.04 & 16.00 & 79.37K \\

AutoLink
& \textbf{98.89} & \textbf{99.76} & 11.39 & 19.28 & 51.81K
& & 73.11 & 82.37 & 8.00 & 12.97 & \textbf{77.93K} \\

DSR-SQL
& 79.40 & 88.14 & 47.19 & 58.58 & 40.01K
& & 72.92 & 80.88 & 21.95 & 30.47 & 281.08K \\

APEX-SQL
& 90.94 & 97.33 & 41.97 & 55.27 & 50.95K
& & 80.85 & 95.26 & 47.50 & 58.64 & 580.80K \\

\midrule
\textbf{EviLink}
& 93.22 & 98.05 & 51.58 & 63.60 & 25.68K
& & \textbf{93.04} & \textbf{97.86} & 26.04 & 36.65 & 116.55K \\

\bottomrule
\end{tabular}
}

\end{table*}

%% file: tables/ablation.tex
\begin{table*}[t]
\caption{Field-level ablation results on BIRD-Dev and Spider2-Snow. Full denotes the EviLink model. Multi and UGAR denote multi-hypothesis grounding and uncertainty-guided agentic refinement, respectively. Static L0--L3 replace adaptive evidence acquisition with fixed evidence levels. SRR, NSR, NSP, and NSF are reported as percentages. Token denotes the average per-example token consumption during schema linking. Higher values are better for metrics except Token.}
\label{tab:ablation}
\centering

{
\setlength{\tabcolsep}{3.0pt}
\renewcommand{\arraystretch}{1.05}

\begin{tabular}{@{}l*{5}{c}c*{5}{c}@{}}
\toprule
\multirow{2}{*}{\textbf{Method}}
& \multicolumn{5}{c}{\textbf{BIRD-Dev} ($N=1534$)}
&
& \multicolumn{5}{c}{\textbf{Spider2-Snow} ($N=517$)} \\
\cmidrule(lr){2-6} \cmidrule(lr){8-12}
& SRR & NSR & NSP & NSF & Token
&
& SRR & NSR & NSP & NSF & Token \\
\midrule

\textbf{Full}
& 92.57 & 97.82 & 56.13 & 67.77 & 28.16K
& & 74.66 & 91.54 & 54.58 & 63.32 & 91.36K \\

\hspace{0.4em}w/o Multi
& 79.92 & 93.61 & 77.89 & 82.59 & 9.03K
& & 63.25 & 86.78 & 60.56 & 66.77 & 44.15K \\

\hspace{0.4em}w/o UGAR
& 92.37 & 97.73 & 55.12 & 66.73 & 25.29K
& & 74.47 & 91.26 & 54.03 & 62.47 & 79.43K \\

\addlinespace[2pt]

Static L0
& 89.18 & 96.91 & 66.17 & 76.10 & 17.56K
& & 62.86 & 88.57 & 64.74 & 71.33 & 76.62K \\

Static L1
& 92.63 & 98.04 & 56.97 & 68.78 & 37.18K
& & 75.24 & 91.33 & 53.19 & 62.16 & 179.63K \\

Static L2
& 91.79 & 97.80 & 61.12 & 72.05 & 52.39K
& & 78.34 & 92.47 & 54.59 & 63.03 & 521.07K \\

Static L3
& 91.66 & 97.66 & 62.55 & 73.22 & 87.45K
& & 74.08 & 92.01 & 55.99 & 63.62 & 1991.14K \\

\bottomrule
\end{tabular}
}

\end{table*}

%% file: tables/sqlgen.tex
\begin{table*}[t]
\caption{Downstream SQL generation performance on BIRD-Dev and Spider2-Snow. Correct denotes the number of correctly answered queries, and EX denotes execution accuracy. Linking and Generation denote the average per-example token consumption during schema linking and SQL generation, respectively. All methods use the same SQL generator, and the best results within each dataset are shown in bold.}
\label{tab:sql_generation_results}
\centering

{
\setlength{\tabcolsep}{3.0pt}

\begin{tabular}{@{}lccccccccc@{}}
\toprule
\multirow{3}{*}{\textbf{Method}}
& \multicolumn{4}{c}{\textbf{BIRD-Dev} ($N=1534$)}
&
& \multicolumn{4}{c}{\textbf{Spider2-Snow} ($N=517$)} \\
\cmidrule(lr){2-5} \cmidrule(lr){7-10}

& \multirow{2}{*}{Correct $\uparrow$}
& \multirow{2}{*}{EX $\uparrow$}
& \multicolumn{2}{c}{Token $\downarrow$}
&
& \multirow{2}{*}{Correct $\uparrow$}
& \multirow{2}{*}{EX $\uparrow$}
& \multicolumn{2}{c}{Token $\downarrow$} \\
\cmidrule(lr){4-5} \cmidrule(lr){9-10}

& & & Linking & Generation
& & & & Linking & Generation \\
\midrule

TA-SQL
& 964 & 62.84 & \textbf{3.38K} & \textbf{2.27K}
& & 157 & 30.37 & 157.41K & \textbf{40.33K} \\

RSL-SQL
& 1004 & 65.45 & 6.13K & 2.80K
& & 190 & 36.75 & 240.34K & 52.41K \\

ReFoRCE
& 450 & 29.34 & 7.29K & 4.18K
& & 169 & 32.69 & 422.47K & 52.64K \\

LinkAlign
& 990 & 64.54 & 8.09K & 9.07K
& & 159 & 30.75 & 79.37K & 72.40K \\

AutoLink
& 1005 & 65.51 & 51.81K & 7.50K
& & 170 & 32.88 & \textbf{77.93K} & 65.82K \\

DSR-SQL
& 985 & 64.21 & 40.01K & 2.99K
& & 171 & 33.08 & 281.08K & 56.08K \\

APEX-SQL
& 1009 & 65.78 & 50.95K & 2.90K
& & 199 & 38.49 & 580.80K & 45.53K \\

\midrule
\textbf{EviLink}
& \textbf{1016} & \textbf{66.23} & 25.68K & 2.83K
& & \textbf{202} & \textbf{39.07} & 116.55K & 50.17K \\

\bottomrule
\end{tabular}
}

\end{table*}

%% file: sections/5_conclusion.tex
\section{Conclusion}

This paper reframes schema linking for large-scale Text-to-SQL along three dimensions. It moves from single-path matching to schema-need inference over multiple plausible SQL paths, from deterministic keep/drop selection to explicit modeling of required and uncertain schema items, and from static evidence provision to uncertainty-guided evidence acquisition. EviLink instantiates this reframing and shows that schema linking should provide not only a compact schema context, but a context that preserves the evidence needed for robust SQL reasoning. Experiments on BIRD-Dev and Spider2-Snow demonstrate that this perspective improves the balance among schema completeness, schema relevance, and token cost, with further gains in downstream SQL generation. These results suggest that robust schema linking can benefit from uncertainty-aware schema-need inference across multiple SQL paths, supported by evidence acquired where needed.

%% file: appendix/baseline.tex
\section{Benchmarks and Baseline Methods}
\label{app:benchmarks_baselines}

This appendix describes the two evaluation benchmarks and the published designs of the seven baselines. We emphasize how each method identifies or explores relevant schema elements, while also describing its original SQL-generation pipeline where necessary. Our experiments evaluate the resulting schema contexts under a shared evaluation protocol, as clarified at the end of this appendix.

\subsection{Benchmarks}
\label{app:benchmark_details}

Following established Text-to-SQL evaluation practice, we select BIRD-Dev and Spider2-Snow as complementary benchmarks. BIRD-Dev evaluates schema linking under realistic database contents and external knowledge requirements, while Spider2-Snow emphasizes enterprise-scale schemas and complex SQL workflows. Together, they allow us to assess schema completeness, relevance, and token cost across different database scales and task complexities.

\paragraph{BIRD-Dev.}
BIRD~\citep{NEURIPS2023_83fc8fab} evaluates Text-to-SQL grounded in realistic database contents. The original benchmark contains 12751 question--SQL pairs over 95 databases spanning 37 domains, with a total database size of 33.4\,GB. Its central challenges include noisy database values, external knowledge required to connect questions to stored data, and efficient query execution. We use its development split, which contains 1534 examples over 11 databases. The released resources include SQLite databases, database descriptions, question-specific evidence, and reference SQL. The benchmark evaluates execution accuracy and also introduces Valid Efficiency Score, which accounts for query efficiency conditional on correctness. In this work, we use BIRD-Dev to examine schema selection under realistic value and knowledge requirements. Its released reference SQL supports the construction of question-specific table and field labels using the procedure in Appendix~\ref{app:gold}; these labels serve as evaluation targets rather than inference inputs.

\paragraph{Spider2-Snow.}
Spider2-Snow is the Snowflake-based Text-to-SQL setting of Spider~2.0~\citep{ICLR2025_46c10f6c}, with 547 tasks accompanied by database metadata and documentation. It is distinct from Spider2-Lite, which spans multiple database systems, and from the broader project-level workflow setting. The original paper reports 152 databases and an average of 812.1 columns per database for Spider2-Snow; these statistics characterize the original benchmark, rather than our retained evaluation subset. Large schemas, nested data structures, domain-specific documentation, and dialect-dependent SQL operations make it a useful setting for evaluating schema linking at enterprise scale. The benchmark uses execution-based evaluation with task-specific answer-comparison criteria. In the snapshot used in our study, reference SQL is released for 120 of the 547 tasks. Following the construction and validation procedure in Appendix~\ref{app:gold}, we obtain reference SQL and schema labels for 517 tasks and evaluate all methods on this same subset. These results therefore characterize the retained subset and are not directly interchangeable with scores on the full official benchmark.

\subsection{Baseline Methods}
\label{app:baseline_details}

\paragraph{TA-SQL.}
TA-SQL~\citep{qu-etal-2024-generation} uses task alignment to reduce hallucinations in schema linking and logical synthesis. Its Task-Aligned Schema Linking module first generates a \emph{dummy SQL query} from the question and database schema, then extracts the referenced schema entities. The dummy query supplies a concrete SQL-based context for selecting tables and columns; it need not already be a correct executable answer. The original framework subsequently applies Task-Aligned Logical Synthesis, representing the required computation through pandas-like symbolic operations whose functions resemble SQL keywords, before producing SQL. The original implementation uses zero-shot prompting for schema linking and few-shot prompting for logical synthesis. The linking component is relevant to our comparison because it grounds schema selection in an intermediate SQL draft, allowing us to examine the coverage and relevance of schema contexts obtained in this way.

\paragraph{RSL-SQL.}
RSL-SQL~\citep{cao2024rslsqlrobustschemalinking} combines forward and backward schema linking to reduce the risk of omitting necessary elements. Forward linking directly selects tables and columns from the question and full schema, with value samples and optional additional context. Backward linking recovers schema elements referenced by a preliminary SQL query. The two selections are united to construct a reduced schema. The paper's chosen backward-linking implementation uses exact column-name matching, favoring recall even when this includes additional same-named columns; it discusses SQLGlot-based extraction as an alternative with a different recall--noise trade-off. Beyond linking, the original system augments the reduced context with descriptions and predicted SQL components, selects between queries generated from full and reduced schemas using their execution results, and applies iterative correction. Thus, its schema-linking contribution is the complementary combination of direct selection and SQL-derived retrieval, while its native query-selection machinery belongs to the broader end-to-end system.

\paragraph{ReFoRCE.}
ReFoRCE~\citep{deng2025reforcetexttosqlagentselfrefinement} addresses enterprise Text-to-SQL through database information compression, SQL self-refinement, consensus enforcement, and column exploration. Compression groups tables with similar naming patterns, retaining detailed metadata for a representative table and names for the remaining tables. When the compressed context remains large, an LLM-based table-selection step further reduces the input. The system generates multiple SQL candidates, refines them with execution feedback, and selects high-confidence answers by agreement among execution results. Cases without a unique leading result are deferred to column exploration, where targeted SQL queries investigate relevant values, nested structures, and dialect-dependent operations before answer generation continues. Exploration queries can themselves be corrected using execution feedback. ReFoRCE is therefore relevant as a system that combines schema reduction with selective, data-grounded exploration.

\paragraph{LinkAlign.}
LinkAlign~\citep{wang-etal-2025-linkalign} separates database retrieval from fine-grained schema grounding in large, potentially multi-database environments. It first retrieves candidate schema information and uses LLM-guided query rewriting to recover potentially missing elements over multiple rounds. It then filters irrelevant candidates and, when multiple databases are present, localizes the target database. Finally, schema parsing identifies the tables and columns needed for SQL generation. The paper provides both a pipeline strategy using a single LLM call for each step and a multi-agent collaboration strategy. In the latter, schema auditing guides query rewriting, while agents with complementary roles assess candidate databases and verify schema selections. For single-database settings, the paper additionally describes random preservation with exponential decay and post-retrieval to mitigate excessive pruning. LinkAlign is relevant to our study as a retrieval-and-reasoning approach to schema grounding.

\paragraph{AutoLink.}
AutoLink~\citep{Wang_Zheng_Cao_Zhang_Wei_Fu_Luo_Chen_Bai_2026} formulates schema linking as iterative exploration and expansion by an LLM agent. The agent interacts with a live database for SQL-based inspection and a vector index for semantic column retrieval. Its initial context contains the question, a complete table-name list, and an initial retrieved schema subset, rather than the full field-level schema. During interaction, the agent can explore schema metadata or data, retrieve additional columns, test schema sufficiency through SQL execution, explicitly add discovered elements, and terminate. The linked schema grows through set union with the added elements, emphasizing recovery of missing information. The original downstream pipeline samples SQL candidates, corrects them using execution feedback, and votes over their execution results. AutoLink provides a direct comparison with agent-driven schema acquisition, particularly the recall--relevance trade-off of progressively expanding a retrieved context.

\paragraph{DSR-SQL.}
DSR-SQL~\citep{hao2025texttosqldualstatereasoningintegrating} organizes Text-to-SQL around a context state and a generation state. Context construction first consolidates structurally similar tables, removes uninformative columns, and summarizes external knowledge. It then selects relevant schema information adaptively: global exploration considers potential SQL structures and tables appearing across high-confidence candidates when the schema fits the context window, whereas partitioned exploration assesses tables individually for larger schemas. A subsequent alignment stage executes lightweight probing queries and summarizes their results into semantic cues such as enumerations and units. The original generation stage progressively extends or revises partial SQL, explores the database when feedback is insufficient, and finalizes the query when its subgoals are resolved. Multiple SQL candidates thus contribute to table selection, after which schema-aware alignment and generation operate on a consolidated schema context. EviLink additionally uses explicit path-wise table and field selections to estimate cross-path support and derive schema-item uncertainty for evidence acquisition.

\paragraph{APEX-SQL.}
APEX-SQL~\citep{cao2026apexsqltalkingdataagentic} grounds schema linking in logical planning and database exploration. It first generates multiple schema-agnostic logical plans and synthesizes them into a unified master plan $P^*$ before schema pruning and grounding. This consolidated plan guides subsequent pruning and semantic-role assignment. Dual-pathway pruning processes schema batches with complementary deletion and preservation passes: a column survives if it is not deleted or is explicitly selected for preservation. Next, semantic linking assigns tentative roles to schema elements, and table-specific agents issue profiling queries to verify these roles against actual data. Global synthesis combines the retained schema and observations into the final schema subgraph. The original SQL-generation stage additionally retrieves exploration guidance and iterates through profiling, consolidation, SQL synthesis, and confirmation. Its multiple initial plans therefore inform exploration through a shared planning context, and the profiling agents are organized by table. EviLink grounds schema needs separately for each path before using cross-path support to construct the required and uncertain sets that guide refinement.

\subsection{Scope of Our Comparison}
\label{app:baseline_comparison_scope}

Our comparison targets schema-context quality under a controlled downstream setting. As described in the main experimental setup, all methods use the same backbone within each benchmark and are evaluated against the same table and field labels. For downstream evaluation, each method supplies its linked schema to the same SQL generator, with the question, prompt templates, and database held fixed. The resulting execution scores measure the utility of the supplied schema contexts under that generator; they do not reproduce the complete native SQL-generation and answer-selection pipelines of all seven systems. Likewise, differences from the original papers' reported scores may reflect the backbone, evaluation subset, schema labels, and generation protocol used here.

Schema-label agreement and downstream execution evaluate complementary properties. The labels in Appendix~\ref{app:gold} describe the tables and fields used by a selected reference SQL query, while another correct query may use a different schema subset. We therefore interpret schema recall and precision relative to these reference labels, and assess answer correctness separately through execution. This distinction is particularly relevant to methods that retain additional schema elements to support alternative SQL realizations.

%% file: appendix/gold.tex
\section{Construction of Gold Tables and Fields}
\label{app:gold}

Spider2-Snow provides gold SQL for only 120 of the 547 examples in our snapshot, limiting the coverage of schema-linking evaluation. Neither Spider2-Snow nor BIRD-Dev directly provides question-specific gold tables and fields.

For BIRD-Dev, we use table and field labels derived from the official SQL for all 1,534 development examples. Tables are identified as \texttt{database.table}, and fields are associated with their parent tables. We verify SQL and database consistency across the official files and check example alignment, including questions and external evidence. We release the reference SQL and evaluation labels.

For Spider2-Snow, we additionally construct and validate reference SQL, yielding 517 annotated examples (94.5\% coverage). Both benchmarks share the schema-label extraction and checking procedure described below; we report detailed Spider2-Snow records to illustrate this procedure. Table~\ref{tab:gold_construction_summary} summarizes the construction and audit records.

\input{tables/gold}

\subsection{Constructing and Auditing Reference SQL}
\label{app:gold_sql_construction}
\label{app:gold_sql_audit}

We use Codex to attempt SQL construction for all 547 examples, including those with released SQL. Given the question, schema metadata, task knowledge, and official answer CSV files, the agent develops Snowflake SQL using database inspection and execution feedback. We use \texttt{GPT-5.6 terra high} for initial construction and \texttt{GPT-5.6 sol high} for retries and semantic repair, without a fixed iteration cap. Instructions prohibit answer enumeration, hardcoding data-derived values or identifiers, and replacing required lookups with answer-derived or database-derived constants. Question-specified constants, unit conversions, output labels, and ordinary null defaults are permitted.

Candidates are executed on Snowflake and checked with the official evaluation functions, respecting designated columns, row order, a floating-point tolerance of 0.01, and alternative answer files. Before semantic repair, the selected references comprise 110 released and 407 constructed queries, with released SQL preferred when both pass. The 30 excluded examples comprise 18 timeout or availability-related failures and 12 execution errors or result mismatches.

We use Claude Code, Codex, and CodeBuddy to review SQL against the question and task knowledge, checking for hardcoding, omitted lookups, and unnecessary dependencies. Review covers joins, filters, existence tests, and row multiplicity, including tables that contribute no output columns. Human review covers 44 examples; these and 15 additional cases identified through other audit signals undergo targeted regeneration. Revisions must pass execution comparison before acceptance; if no accepted revision is produced, the existing reference is retained. Final SQL-audit records contain no outstanding flags for the 517 retained queries.

Of the 110 released queries selected before semantic repair, 15 are subsequently revised, leaving 95 unchanged official references in the final set. Appendix~\ref{app:official_sql_subset} reports results on this subset.

\subsection{Extracting and Checking Schema Labels}
\label{app:gold_schema_extraction}

For both benchmarks, table labels identify fully qualified physical tables, and field labels identify physical-table/column pairs. We use SQLGlot with the corresponding SQL dialect to resolve aliases and column lineage across nested query structures. The Spider2-Snow records reported here use SQLGlot 29.0.1 in the Snowflake dialect. Extraction includes references in joins, filters, grouping, ordering, \texttt{HAVING}, and \texttt{QUALIFY}. Catalog metadata supports name resolution without excluding valid cross-database references. Table~\ref{tab:gold_annotation_rules} specifies the annotation conventions.

\input{tables/schema_gold}

Field labels follow an explicit-reference convention: direct \texttt{SELECT *} does not by itself add all output columns, although these remain execution dependencies. Field labels therefore do not represent complete execution lineage. All retained examples have nonempty field labels. A separate validator checks label structure, catalog membership, and missing catalog-known table references, while targeted checks correct extraction errors.

One dynamic-SQL example uses a documented static expansion for label extraction while retaining the original SQL as its execution reference. Catalog validation is limited for three SQL-referenced fields absent from the supplied metadata and one annotated table with an empty catalog entry.

Claude Code, Codex, and CodeBuddy check the final labels for missing or extra tables and fields on both benchmarks. In the Spider2-Snow records reported here, all 517 examples receive approving judgments from all three agents. These checks complement extractor corrections and, for Spider2-Snow, the earlier human SQL review; agreement does not provide an independent estimate of annotation accuracy. Records retain final SQL, labels, review comments, and label changes.

\subsection{Evaluation Set and Protocol}
\label{app:gold_final_audit}

The 517 annotated Spider2-Snow examples form our schema-linking evaluation set. We correct two database identifiers to match the databases referenced by their SQL, leaving questions, example membership, and schema labels unchanged. All methods use the same examples, labels, and corrected metadata, with reference SQL, answer files, and schema labels excluded from inference inputs.

Schema metrics measure agreement with one selected reference SQL path per question, while downstream SQL is evaluated separately against official answers. Correct alternatives may use different tables or fields, and execution agreement on the available database does not establish equivalence across all database instances. Results apply to the retained 517 examples, with potential selection bias toward tasks whose reference SQL can be constructed in the available environment.

\subsection{Evaluation on the Official-SQL Subset}
\label{app:official_sql_subset}

We evaluate a subset with unchanged officially released SQL to examine whether the main findings also hold without constructed or revised reference queries. Of the 120 examples with released SQL in our snapshot, 113 belong to the final 517-example evaluation set. Among these, 18 use final reference SQL that differs from the official text, leaving 95 examples whose final references are byte-identical to the released SQL. The SQL used for schema-label extraction also matches the official text for all 95 examples. Subset membership is determined by reference provenance and exact SQL agreement, not method performance.

We reuse the frozen schema-linking and downstream SQL predictions from the main experiments, restricting evaluation and token aggregation to the same 95 examples for every method. Schema linking uses GLM-5.1, and downstream SQL generation uses Qwen3.6-27B with the same generation protocol across methods. Schema labels follow the extraction and validation procedures above, and downstream EX is evaluated against the official answers.

\input{tables/official_schema}

Table~\ref{tab:official_sql_schema_linking} shows that EviLink achieves the highest field-level SRR and NSR, at 91.58\% and 98.17\%, respectively, with the lowest linking token consumption of 55.27K per example. Compared with APEX-SQL, it uses 84.8\% fewer linking tokens while obtaining higher field-level recall, although its NSP and NSF remain lower. At the table level, EviLink obtains 98.95\% SRR and 99.91\% NSR. RSL-SQL reaches 100\% on both recall metrics, but EviLink retains higher table precision, with 56.57\% NSP versus 28.66\%.

\input{tables/official_sql}

Table~\ref{tab:official_sql_generation} reports downstream performance on the same subset. EviLink achieves the highest EX at 43.16\% (41/95), compared with 42.11\% (40/95) for APEX-SQL. The downstream gain is one query, while generation token consumption is higher than APEX-SQL's (47.28K vs.\ 39.03K). These results support the main findings on field-level schema completeness and linking efficiency using unchanged official reference SQL.

%% file: tables/gold.tex
\begin{table}[t]
\caption{Spider2-Snow annotation statistics. SQL sources are counted before semantic repair. Audit categories overlap. Table and field totals count label instances across examples.}
\label{tab:gold_construction_summary}
\centering
\small
\setlength{\tabcolsep}{5pt}
\renewcommand{\arraystretch}{1.05}
\begin{tabular}{@{}p{0.81\linewidth}r@{}}
\toprule
\textbf{Statistic} & \textbf{Count} \\
\midrule
Accepted / excluded examples, out of 547 & 517 / 30 \\
Released / constructed SQL selected at the merge & 110 / 407 \\
Recorded construction attempts & 1079 \\
\midrule
Examples flagged in the initial SQL audit & 49 \\
Examples with conflicting initial judgments & 39 \\
Examples with documented human SQL review & 44 \\
Examples entering targeted regeneration & 59 \\
\midrule
Examples with changed table / field labels after refinement & 4 / 30 \\
Final table / field instances & 3770 / 14324 \\
Examples approved by all three final label reviews & 517 \\
Human adjudications in the final label review & 0 \\
\bottomrule
\end{tabular}
\end{table}

%% file: tables/schema_gold.tex
\begin{table}[t]
\caption{Schema annotation rules. Field labels follow explicit top-level column references after scope and lineage resolution.}
\label{tab:gold_annotation_rules}
\centering
\small
\setlength{\tabcolsep}{4pt}
\renewcommand{\arraystretch}{1.08}
\begin{tabular}{@{}p{0.26\linewidth}p{0.68\linewidth}@{}}
\toprule
\textbf{Construct} & \textbf{Rule} \\
\midrule
CTEs and aliases & Trace references to physical sources; exclude derived names. \\
Correlated columns & Resolve the source in the outer scope defining its alias. \\
Nested fields and \texttt{FLATTEN} & Label the top-level physical column; distinguish pseudo-columns from real columns with the same names. \\
\texttt{COUNT(*)} & Label the source table without adding every column. \\
Intermediate \texttt{SELECT *} & Trace columns explicitly referenced downstream. \\
Direct \texttt{SELECT *} & Do not expand the wildcard to all catalog columns. \\
Table shards & Label each referenced physical shard separately. \\
\bottomrule
\end{tabular}
\end{table}

%% file: tables/official_schema.tex
\begin{table*}[t]
\caption{Schema-linking performance on the Spider2-Snow subset with unchanged officially released SQL ($N=95$), using GLM-5.1. Token denotes the average per-example token consumption during schema linking; the same cost is reported at both evaluation levels. Higher values are better for all metrics except Token.}
\label{tab:official_sql_schema_linking}
\centering

{
\setlength{\tabcolsep}{4.0pt}

\begin{tabular}{@{}lccccccccccc@{}}
\toprule
\multirow{2}{*}{\textbf{Method}}
& \multicolumn{5}{c}{\textbf{Field level} ($N=95$)}
&
& \multicolumn{5}{c}{\textbf{Table level} ($N=95$)} \\
\cmidrule(lr){2-6} \cmidrule(lr){8-12}
& SRR & NSR & NSP & NSF & Token
&
& SRR & NSR & NSP & NSF & Token \\
\midrule

TA-SQL
& 40.00 & 76.55 & \textbf{75.36} & \textbf{73.15} & 70.41K
& & 77.89 & 89.09 & \textbf{89.23} & \textbf{87.03} & 70.41K \\

RSL-SQL
& 75.79 & 95.83 & 24.06 & 35.30 & 224.75K
& & \textbf{100.00} & \textbf{100.00} & 28.66 & 39.47 & 224.75K \\

ReFoRCE
& 35.79 & 83.16 & 54.20 & 59.87 & 165.30K
& & 93.68 & 98.07 & 59.67 & 67.86 & 165.30K \\

LinkAlign
& 68.42 & 83.65 & 8.73 & 15.12 & 56.94K
& & 94.74 & 95.79 & 23.41 & 34.67 & 56.94K \\

AutoLink
& 85.26 & 93.24 & 8.52 & 14.87 & 72.78K
& & 91.58 & 95.61 & 23.63 & 35.16 & 72.78K \\

DSR-SQL
& 89.47 & 93.70 & 27.14 & 37.40 & 169.33K
& & 89.47 & 93.80 & 78.25 & 82.94 & 169.33K \\

APEX-SQL
& 82.11 & 96.14 & 47.88 & 59.12 & 362.43K
& & 97.89 & 99.39 & 50.60 & 60.95 & 362.43K \\

\midrule
\textbf{EviLink}
& \textbf{91.58} & \textbf{98.17} & 29.20 & 40.17 & \textbf{55.27K}
& & 98.95 & 99.91 & 56.57 & 66.42 & \textbf{55.27K} \\

\bottomrule
\end{tabular}
}

\end{table*}

%% file: tables/official_sql.tex
\begin{table}[t]
\caption{Downstream SQL generation performance on the same official-SQL subset ($N=95$). All methods use GLM-5.1 for schema linking and Qwen3.6-27B for SQL generation. Correct denotes the number of correctly answered queries, and EX denotes execution accuracy. Linking and Generation denote the average per-example token consumption in the corresponding stages.}
\label{tab:official_sql_generation}
\centering

{
\setlength{\tabcolsep}{14.0pt}

\begin{tabular}{@{}lcccc@{}}
\toprule
\multirow{2}{*}{\textbf{Method}}
& \multirow{2}{*}{Correct $\uparrow$}
& \multirow{2}{*}{EX (\%) $\uparrow$}
& \multicolumn{2}{c}{\textbf{Token} $\downarrow$} \\
\cmidrule(lr){4-5}
& & & Linking & Generation \\
\midrule

TA-SQL
& 28 & 29.47 & 70.41K & 40.49K \\

RSL-SQL
& 35 & 36.84 & 224.75K & 49.21K \\

ReFoRCE
& 36 & 37.89 & 165.30K & 45.66K \\

LinkAlign
& 32 & 33.68 & 56.94K & 71.40K \\

AutoLink
& 31 & 32.63 & 72.78K & 69.70K \\

DSR-SQL
& 34 & 35.79 & 169.33K & 55.55K \\

APEX-SQL
& 40 & 42.11 & 362.43K & \textbf{39.03K} \\

\midrule
\textbf{EviLink}
& \textbf{41} & \textbf{43.16} & \textbf{55.27K} & 47.28K \\

\bottomrule
\end{tabular}
}

\end{table}

%% file: appendix/metric.tex
\section{Metric Computation}
\label{app:metric}

This appendix provides the computation details for the schema-linking metrics introduced in Section~\ref{sec:experimental_setup}. For each evaluable question \(i\), let \(G_i\) denote the gold schema item set and \(\hat{G}_i\) denote the predicted schema item set. The item unit can be either a table or a field, depending on the evaluation granularity.

For each question, we compute:
\begin{equation}
\mathrm{TP}_i = |G_i \cap \hat{G}_i|.
\end{equation}
\begin{equation}
n_i^G = |G_i|, \quad n_i^P = |\hat{G}_i|.
\end{equation}
\begin{equation}
R_i = \frac{\mathrm{TP}_i}{n_i^G}, \quad
P_i = \frac{\mathrm{TP}_i}{n_i^P}.
\end{equation}
\begin{equation}
F_{1,i} = \frac{2P_iR_i}{P_i + R_i}.
\end{equation}
If a denominator is zero, the corresponding score is set to zero. Empty predictions are included in the evaluation set.

We define \(s_i = 1\) if \(R_i = 1\), and \(s_i = 0\) otherwise. Given \(N\) evaluable questions, the reported metrics are:
\begin{equation}
\mathrm{SRR} = \frac{1}{N}\sum_{i=1}^{N} s_i, \quad
\mathrm{NSR} = \frac{1}{N}\sum_{i=1}^{N} R_i.
\end{equation}
\begin{equation}
\mathrm{NSP} = \frac{1}{N}\sum_{i=1}^{N} P_i, \quad
\mathrm{NSF} = \frac{1}{N}\sum_{i=1}^{N} F_{1,i}.
\end{equation}

The same computation is applied to table-level and field-level evaluation by changing the schema item unit. In the main experiments, we report field-level results by default.

%% file: appendix/settings.tex
\section{Parameter Settings}
\label{app:parameter}

\input{tables/model_para}

\paragraph{Model assignment and inference configurations.}
Table~\ref{tab:method_parameters} summarizes the inference configurations for each experiment.
For the main schema-linking comparisons, all methods use Qwen3.6-35B-A3B on BIRD-Dev and GLM-5.1 on Spider2-Snow, while retaining their method-specific procedures.
The schema-linking ablations and hypothesis-budget analysis use Qwen3.8-27B.
Across these experiments, EviLink uses temperature $1.0$, top-$p=0.95$, and a maximum generation length of $16384$ tokens per call, with reasoning requested to be disabled.
All token limits in this section refer to configured maximum generation lengths per call, not context-window sizes or measured token consumption.

\paragraph{Additional backbone evaluation.}
We additionally evaluate EviLink on Spider2-Snow with GLM-5.2, DeepSeek-4 Pro, GLM-5.3, and Qwen3.8-27B.
All four configurations use temperature $1.0$, top-$p=0.95$, and a maximum generation length of $65536$ tokens per call.
We request low-effort reasoning for GLM-5.3 and disabled reasoning for the other three backbones.
This evaluation therefore compares the listed backbone configurations, including their respective reasoning settings.

\paragraph{Downstream SQL generation.}
We use Qwen3.8-27B on BIRD-Dev and Qwen3.6-27B on Spider2-Snow.
Within each dataset, all compared methods share the same SQL generator and candidate-generation, revision, and selection procedure, using their respective schema-linking outputs.
Both protocols use temperature $0.7$, top-$p=0.8$, top-$k=20$, minimum sampling probability $0$, presence penalty $1.5$, and repetition penalty $1.0$, with reasoning requested to be disabled.
They generate three SQL candidates and allow up to three revision rounds, using a base seed of $42$.
The maximum generation length is set to $65536$ tokens per call on both datasets. The protocols account for the database dialect and evaluation procedure of each benchmark.
Reference schema labels and gold execution results are not provided to schema linking or SQL candidate selection during inference.

\input{tables/parameters}

\paragraph{Default EviLink method settings.}
Table~\ref{tab:algorithm_settings} summarizes the default method parameters.
Multi-hypothesis grounding generates at most $K=4$ SQL hypotheses and may return fewer when the model considers them sufficient.
Table selection starts with L0 evidence, field selection uses L1 evidence, and refinement uses L2 evidence by default, acquiring L3 evidence on demand.
The required-set threshold is $\tau_{\mathrm{req}}=0.85$.
An auxiliary threshold of $0.50$ partitions uncertain items during refinement; it does not determine the final keep/discard decision.
Refinement has separate budgets of up to eight turns for uncertain-item processing and up to three turns for candidate recovery.
Both budgets are upper bounds, and the corresponding procedures may terminate earlier.

\paragraph{Ablation and sensitivity settings.}
Within each dataset, all schema-linking ablation variants use the same backbone, inference configuration, and evaluation set.
The single-hypothesis variant sets $K=1$, while the variant without uncertainty-guided agentic refinement (UGAR) directly uses the grounding output without refinement.
Static-L0 through Static-L3 each use a fixed evidence level throughout table selection, field selection, and refinement.
The hypothesis-budget analysis varies $K\in\{1,\ldots,6\}$ on a fixed subset of $200$ instances randomly sampled from our Spider2-Snow evaluation set ($N=517$) using seed $42$.
All other method settings remain unchanged in this analysis.
Dataset-specific prompts and configuration files accompany our implementation.

%% file: tables/model_para.tex
\begin{table}[t]
\caption{Experiment-specific inference configurations.}
\label{tab:method_parameters}
\centering
\begin{tabular}{llccrc}
\toprule
\textbf{Dataset}
& \textbf{Model}
& \textbf{Temp.}
& \textbf{Top-$p$}
& \textbf{Token limit}
& \textbf{Reasoning} \\
\midrule
\multicolumn{6}{l}{\textit{Main schema linking}} \\
BIRD-Dev & Qwen3.6-35B-A3B
& 1.0 & 0.95 & 16384 & Off \\
Spider2-Snow & GLM-5.1
& 1.0 & 0.95 & 16384 & Off \\
\midrule
\multicolumn{6}{l}{\textit{Schema-linking ablation}} \\
Both & Qwen3.8-27B
& 1.0 & 0.95 & 16384 & Off \\
\midrule
\multicolumn{6}{l}{\textit{Hypothesis-budget analysis}} \\
Spider2-Snow & Qwen3.8-27B
& 1.0 & 0.95 & 16384 & Off \\
\midrule
\multicolumn{6}{l}{\textit{Main SQL generation}} \\
BIRD-Dev & Qwen3.8-27B
& 0.7 & 0.8 & 65536 & Off \\
Spider2-Snow & Qwen3.6-27B
& 0.7 & 0.8 & 65536 & Off \\
\midrule
\multicolumn{6}{l}{\textit{Backbone comparison}} \\
Spider2-Snow & GLM-5.2
& 1.0 & 0.95 & 65536 & Off \\
Spider2-Snow & DeepSeek-4 Pro
& 1.0 & 0.95 & 65536 & Off \\
Spider2-Snow & GLM-5.3
& 1.0 & 0.95 & 65536 & Low \\
Spider2-Snow & Qwen3.8-27B
& 1.0 & 0.95 & 65536 & Off \\
\bottomrule
\end{tabular}
\end{table}

%% file: tables/parameters.tex
\begin{table}[t]
\caption{Default EviLink method parameters and downstream SQL-generation budgets.}
\label{tab:algorithm_settings}
\centering

\begin{tabular}{lr}
\toprule
\textbf{Parameter} & \textbf{Value} \\
\midrule

\multicolumn{2}{l}{\textit{Multi-hypothesis grounding}} \\
Maximum hypotheses, $K$ & 4 \\
Hypothesis count & Adaptive ($1$--$4$) \\
Table evidence & L0 \\
Field evidence & L1 \\

\midrule
\multicolumn{2}{l}{\textit{Agentic refinement}} \\
Base evidence & L2 \\
On-demand evidence & L3 \\
Uncertain-item turn budget & 8 \\
Candidate-recovery turn budget & 3 \\

\midrule
\multicolumn{2}{l}{\textit{Uncertainty thresholds}} \\
Required threshold, $\tau_{\mathrm{req}}$ & 0.85 \\
Uncertain-item partition threshold & 0.50 \\

\midrule
\multicolumn{2}{l}{\textit{Main SQL generation}} \\
SQL candidates & 3 \\
Maximum revision rounds & 3 \\
Base seed & 42 \\

\bottomrule
\end{tabular}
\end{table}

%% file: appendix/table_results.tex
\section{Table-Level Results}
\label{app:table_results}

\paragraph{Main results.}
Table~\ref{tab:table_schema_linking_results} shows that EviLink achieves high table recall while retaining higher precision than strongly recall-oriented baselines. On BIRD-Dev, EviLink obtains 96.94\% SRR and 98.79\% NSR, together with 75.67\% NSP and 83.12\% NSF, using 25.68K tokens per example. Although RSL-SQL, LinkAlign, AutoLink, and APEX-SQL achieve higher recall, their NSP ranges from 33.06\% to 45.95\%, indicating a greater proportion of irrelevant tables in their predictions. On Spider2-Snow, EviLink achieves the highest SRR and NSR at 96.91\% and 98.64\%, respectively. It also exceeds RSL-SQL and APEX-SQL in NSP and NSF, while using 116.55K tokens compared with 240.34K and 580.80K, respectively. TA-SQL achieves higher NSP and NSF on both benchmarks, but lower recall, particularly on Spider2-Snow, where its SRR is 68.86\%. These results highlight EviLink's ability to retain necessary tables while limiting irrelevant table inclusion relative to other high-recall methods.

\input{tables/mainstream_table}
\input{tables/ablation_table}

\paragraph{Ablation study.}
Table~\ref{tab:ablation_table_level} examines the contributions of multi-hypothesis grounding, agentic refinement, and adaptive evidence acquisition using Qwen3.8-27B on both benchmarks. Removing multi-hypothesis grounding reduces SRR from 96.48\% to 92.50\% on BIRD-Dev and from 88.59\% to 82.59\% on Spider2-Snow, while increasing precision and reducing token consumption. This trade-off supports its role in recovering necessary tables that a single hypothesis may miss. Refinement provides smaller recall gains: relative to w/o UGAR, Full increases SRR by 0.20 and 0.97 percentage points on the two benchmarks, respectively, with little change in NSP or NSF. Compared with Static L1--L3, adaptive evidence acquisition uses fewer tokens on both benchmarks. On Spider2-Snow, Full matches Static L1's SRR with 91.36K rather than 179.63K tokens, whereas Static L2 reaches a higher SRR of 90.91\% at 521.07K tokens. Static L0 is cheaper and yields higher NSP and NSF, but lower SRR on both benchmarks. These results indicate that multi-hypothesis grounding improves table coverage, while adaptive evidence acquisition offers a practical trade-off between coverage and token consumption.

%% file: tables/mainstream_table.tex
\begin{table*}[t]
\caption{Table-level schema linking performance on BIRD-Dev and Spider2-Snow. Token denotes the average per-example token consumption during schema linking. Higher values are better for all metrics except Token. Best results within each dataset are shown in bold.}
\label{tab:table_schema_linking_results}
\centering

{
\setlength{\tabcolsep}{4.0pt}

\begin{tabular}{@{}lccccccccccc@{}}
\toprule
\multirow{2}{*}{\textbf{Method}}
& \multicolumn{5}{c}{\textbf{BIRD-Dev} ($N=1534$)}
&
& \multicolumn{5}{c}{\textbf{Spider2-Snow} ($N=517$)} \\
\cmidrule(lr){2-6} \cmidrule(lr){8-12}
& SRR & NSR & NSP & NSF & Token
&
& SRR & NSR & NSP & NSF & Token \\
\midrule

TA-SQL
& 90.03 & 95.41 & \textbf{92.13} & \textbf{92.66} & \textbf{3.38K}
& & 68.86 & 81.50 & \textbf{82.92} & \textbf{79.13} & 157.41K \\

RSL-SQL
& 98.83 & 99.53 & 45.95 & 59.11 & 6.13K
& & 96.32 & 98.52 & 31.97 & 41.59 & 240.34K \\

ReFoRCE
& 39.63 & 64.42 & 79.90 & 67.83 & 7.29K
& & 89.75 & 95.86 & 55.86 & 63.34 & 422.47K \\

LinkAlign
& \textbf{100.00} & \textbf{100.00} & 33.06 & 47.10 & 8.09K
& & 88.20 & 94.18 & 24.48 & 34.35 & 79.37K \\

AutoLink
& 99.93 & 99.97 & 33.93 & 48.01 & 51.81K
& & 78.92 & 84.87 & 24.27 & 31.81 & \textbf{77.93K} \\

DSR-SQL
& 94.72 & 97.71 & 81.88 & 86.73 & 40.01K
& & 73.11 & 81.28 & 70.67 & 72.35 & 281.08K \\

APEX-SQL
& 99.02 & 99.63 & 41.33 & 54.65 & 50.95K
& & 94.00 & 97.93 & 55.45 & 64.76 & 580.80K \\

\midrule
\textbf{EviLink}
& 96.94 & 98.79 & 75.67 & 83.12 & 25.68K
& & \textbf{96.91} & \textbf{98.64} & 58.85 & 67.05 & 116.55K \\

\bottomrule
\end{tabular}
}

\end{table*}

%% file: tables/ablation_table.tex
\begin{table*}[t]
\caption{Table-level ablation results on BIRD-Dev and Spider2-Snow. Full denotes the complete EviLink model. Multi and UGAR denote multi-hypothesis grounding and uncertainty-guided agentic refinement, respectively. Static L0--L3 replace adaptive evidence acquisition with fixed evidence levels. SRR, NSR, NSP, and NSF are reported as percentages. Token denotes the average per-example token consumption during schema linking. Higher values are better for all metrics except Token.}
\label{tab:ablation_table_level}
\centering

{
\setlength{\tabcolsep}{3.0pt}
\renewcommand{\arraystretch}{1.05}

\begin{tabular}{@{}l*{5}{c}c*{5}{c}@{}}
\toprule
\multirow{2}{*}{\textbf{Method}}
& \multicolumn{5}{c}{\textbf{BIRD-Dev} ($N=1534$)}
&
& \multicolumn{5}{c}{\textbf{Spider2-Snow} ($N=517$)} \\
\cmidrule(lr){2-6} \cmidrule(lr){8-12}
& SRR & NSR & NSP & NSF & Token
&
& SRR & NSR & NSP & NSF & Token \\
\midrule

\textbf{Full}
& 96.48 & 98.55 & 77.78 & 84.69 & 28.16K
& & 88.59 & 94.97 & 78.88 & 82.56 & 91.36K \\

\hspace{0.4em}w/o Multi
& 92.50 & 96.71 & 90.86 & 92.53 & 9.03K
& & 82.59 & 91.90 & 81.86 & 83.42 & 44.15K \\

\hspace{0.4em}w/o UGAR
& 96.28 & 98.47 & 77.85 & 84.71 & 25.29K
& & 87.62 & 94.61 & 79.08 & 82.46 & 79.43K \\

\addlinespace[2pt]

Static L0
& 96.09 & 98.38 & 80.96 & 86.76 & 17.56K
& & 85.11 & 94.28 & 82.82 & 85.76 & 76.62K \\

Static L1
& 96.81 & 98.77 & 78.95 & 85.48 & 37.18K
& & 88.59 & 94.62 & 77.11 & 80.83 & 179.63K \\

Static L2
& 96.35 & 98.56 & 81.26 & 86.94 & 52.39K
& & 90.91 & 95.81 & 77.51 & 81.29 & 521.07K \\

Static L3
& 96.81 & 98.74 & 82.39 & 87.83 & 87.45K
& & 88.97 & 95.18 & 76.31 & 79.64 & 1991.14K \\

\bottomrule
\end{tabular}
}

\end{table*}

%% file: appendix/size.tex
\section{Performance across Schema Sizes}
\label{app:schema_size}

\input{tables/size}

We analyze the frozen GLM-5.1 main-experiment predictions for all 517 Spider2-Snow examples spanning 135 databases. We group examples using fixed thresholds on the number of top-level fields in the full database schema: small ($\leq91$), medium ($92$--$505$), and large ($\geq506$), containing 171, 178, and 168 examples, respectively. Table~\ref{tab:schema_size_results} compares field-level SRR across all eight methods.

On small schemas, EviLink achieves 98.25\% SRR while retaining a more precise schema context than LinkAlign, with field-level NSP of 32.16\% versus 14.14\%. This demonstrates strong schema completeness together with more selective field retention.

EviLink achieves the highest field-level SRR on both medium and large schemas. Its advantage over the strongest baseline in each group increases from 5.06 percentage points over DSR-SQL on medium schemas to 14.29 percentage points over RSL-SQL on large schemas. In the large-schema group, EviLink retains all reference fields on 148 of 168 examples, compared with 124 for RSL-SQL. It also exceeds APEX-SQL across all three groups, with the largest margin of 18.45 percentage points on large schemas. These results show that EviLink's recall advantage is particularly pronounced in the large-schema setting targeted by our method.

EviLink uses an average of 21.56K, 39.21K, and 295.19K linking tokens in the small, medium, and large groups, respectively. On large schemas, its token consumption is 80.85\% lower than APEX-SQL's, combining stronger schema completeness with substantially lower linking cost.

%% file: tables/size.tex
\begin{table}[t]
\caption{Field-level SRR across schema sizes on Spider2-Snow ($N=517$), using GLM-5.1. Groups are defined by the number of top-level fields in the full database schema. SRR is reported as a percentage, and the best result in each group is shown in bold. Gains discussed in the text are computed from unrounded scores.}
\label{tab:schema_size_results}
\centering
{
\setlength{\tabcolsep}{8pt}
\begin{tabular}{@{}lccc@{}}
\toprule
\multirow{2}{*}{\textbf{Method}}
& \textbf{Small} ($N=171$)
& \textbf{Medium} ($N=178$)
& \textbf{Large} ($N=168$) \\
& $\leq91$ fields
& $92$--$505$ fields
& $\geq506$ fields \\
\midrule
TA-SQL
& 49.71 & 43.82 & 35.12 \\
RSL-SQL
& 83.63 & 80.90 & 73.81 \\
ReFoRCE
& 47.95 & 39.89 & 33.33 \\
LinkAlign
& \textbf{100.00} & 74.16 & 14.88 \\
AutoLink
& 99.42 & 87.08 & 31.55 \\
DSR-SQL
& 91.81 & 87.64 & 38.10 \\
APEX-SQL
& 87.72 & 84.83 & 69.64 \\
\midrule
\textbf{EviLink}
& 98.25 & \textbf{92.70} & \textbf{88.10} \\
\bottomrule
\end{tabular}
}
\end{table}

%% file: appendix/more_models.tex
\section{Additional Backbone Evaluation}
\label{app:additional_backbone}

\input{tables/additional_model}

We further evaluate EviLink with four additional LLM backbones (GLM-5.2, GLM-5.3 Low, DeepSeek-4 Pro, and Qwen3.8) on the same 517 Spider2-Snow examples to examine its applicability across models. As shown in Table~\ref{tab:backbone_comparison}, EviLink maintains high schema recall across all four configurations, with field-level NSR ranging from 91.45\% to 98.70\% and table-level NSR from 94.29\% to 98.94\%. Among these configurations, GLM-5.3 Low achieves the highest field- and table-level SRR of 96.52\% and 97.68\%, respectively, indicating strong complete-schema coverage. Qwen3.8 achieves the highest NSF at both the field level (62.72\%) and the table level (81.45\%), together with the lowest token consumption of 87.18K per example. These results support the applicability of EviLink across multiple model families, with high schema recall maintained across the evaluated backbones.

%% file: tables/additional_model.tex
\begin{table*}[t]
\caption{Schema linking performance and token consumption of EviLink with four LLM backbones on Spider2-Snow ($N=517$). SRR, NSR, NSP, and NSF are reported as percentages. Token denotes the average schema linking token consumption per example, in thousands; the same cost is reported at both evaluation levels. Bold indicates the highest schema linking scores and the lowest token consumption within each evaluation level.}
\label{tab:backbone_comparison}
\centering

\begin{tabular}{lccccc}
\toprule
\multirow{2}{*}{\textbf{Backbone}}
& \multicolumn{5}{c}{\textbf{Spider2-Snow} ($N=517$)} \\
\cmidrule{2-6}
& SRR & NSR & NSP & NSF & Token \\
\midrule

\multicolumn{6}{l}{\textit{Field level}} \\

GLM-5.2
& 94.97 & 98.51 & 22.29 & 32.53 & 163.73K \\

DeepSeek-4 Pro
& 92.07 & 97.52 & 26.42 & 36.89 & 170.82K \\

Qwen3.8-27B
& 73.31 & 91.45 & \textbf{53.60} & \textbf{62.72}
& \textbf{87.18K} \\

GLM-5.3 Low
& \textbf{96.52} & \textbf{98.70} & 25.46 & 37.18 & 130.16K \\

\midrule
\multicolumn{6}{l}{\textit{Table level}} \\

GLM-5.2
& 97.49 & \textbf{98.94} & 51.84 & 60.98 & 163.73K \\

DeepSeek-4 Pro
& 95.36 & 98.11 & 61.68 & 69.37 & 170.82K \\

Qwen3.8-27B
& 86.85 & 94.29 & \textbf{77.51} & \textbf{81.45}
& \textbf{87.18K} \\

GLM-5.3 Low
& \textbf{97.68} & 98.86 & 54.77 & 64.58 & 130.16K \\

\bottomrule
\end{tabular}
\end{table*}

%% file: appendix/multiple.tex
\section{Understanding Multiple SQL Paths and Gold Schema Variability}
\label{app:multiple}

\input{tables/multi_gold}

\paragraph{Why multiple SQL paths matter.}
A central premise of EviLink is that a natural-language question in large-scale Text-to-SQL does not always determine a unique SQL realization. In practice, valid solutions may differ in source tables, projection granularity, aggregation units, output organization, or the auxiliary fields used to instantiate a particular query strategy. Since gold tables and fields are derived from SQL dependencies, different SQL realizations can induce different reference schema annotations. We examine this phenomenon from two complementary sources: downstream SQL generation and gold SQL construction.

\paragraph{Evidence from downstream SQL generation.}
In the downstream SQL-generation experiment, we find cases where EviLink has perfect table recall but imperfect field recall, while the final generated SQL is accepted by the execution evaluator. In these cases, the missing gold-labeled fields are not referenced by the selected SQL. For example, Table~\ref{tab:multiple_sql_paths} shows that a Stack Overflow query can return question identifiers and user statistics without projecting the reference title field. A census-tract donation query and a medical-image segmentation query also produce accepted outputs without using the reference entity-type field and image-reference sequence fields, respectively. These cases show that field-level recall is a useful diagnostic, but not a perfect proxy for execution-based acceptance: reference-specific fields need not appear in every accepted SQL realization. Acceptance on the evaluated database, however, does not by itself establish that every omitted field is semantically unnecessary.

\paragraph{Evidence from gold SQL construction.}
We observe a related phenomenon when comparing the released Spider2-Snow SQL with our final constructed reference SQL. Representative examples are shown in Table~\ref{tab:multiple_sql_paths}. An RFM query obtains monetary values from payment records versus item-level prices and freight charges; a source-code analysis query uses language-specific retrieval branches versus a unified file--content join; and a shopping-cart query uses event-sequence attribution versus purchase-visit membership with an event-dictionary join. These differences are not merely surface-level variations: they change the source tables, join structures, or fields consumed by the queries, and can therefore induce different reference schema sets. Some pairs also differ in filtering or aggregation semantics, so implementation differences alone should not be interpreted as proof of semantic equivalence. Nevertheless, they illustrate why schema annotations must be understood in relation to the particular SQL realization from which they are derived.

\paragraph{Takeaway.}
Together, these observations motivate our formulation of schema linking as schema-need inference under multiple plausible SQL hypotheses. A single reference SQL provides a useful evaluation target, but its dependency set should not automatically be equated with the unique set of question-essential schema items. This motivates EviLink to preserve schema evidence for multiple plausible reasoning paths, distinguish required from uncertain schema items, and acquire additional evidence only when uncertainty remains.

%% file: tables/multi_gold.tex
\begin{table*}[t]
\caption{Representative schema-dependency differences in downstream generation and gold SQL construction. In the upper examples, EviLink's linked schemas have complete table recall but incomplete field recall, while the generated SQL is accepted by the evaluator. The lower examples compare released and constructed SQL implementations; implementation differences alone do not establish semantic equivalence.}
\label{tab:multiple_sql_paths}
\centering
\setlength{\tabcolsep}{0pt}
\renewcommand{\arraystretch}{1.16}
\newcommand{\tc}[1]{\parbox[t]{\linewidth}{\raggedright\strut #1\strut}}

\begin{tabular}{@{}p{0.090\textwidth}@{\hspace{12pt}}p{0.135\textwidth}@{\hspace{14pt}}p{0.360\textwidth}@{\hspace{14pt}}p{0.315\textwidth}@{}}
\toprule
\tc{\textbf{Source}} & \tc{\textbf{Case}} & \tc{\textbf{Alternative SQL paths}} & \tc{\textbf{Schema implication}} \\
\midrule

\multicolumn{4}{@{}l}{\textit{Downstream SQL generation: evaluator-accepted SQL despite imperfect field recall}} \\
\midrule

\tc{SQLGen}
& \tc{\mbox{\texttt{sf\_bq309}}}
& \tc{The reference projects \texttt{title}; generated SQL omits it and returns question identifiers, body lengths, and user statistics.}
& \tc{A reference projection can contain fields outside the evaluated output requirements.} \\[5pt]

\tc{SQLGen}
& \tc{\mbox{\texttt{sf\_bq023}}}
& \tc{The reference uses \texttt{entity\_tp} to filter donations; generated SQL omits this filter and computes tract-level averages matching an accepted reference output.}
& \tc{Execution-based acceptance can coexist with omission of a reference filtering field.} \\[5pt]

\tc{SQLGen}
& \tc{\mbox{\texttt{sf\_bq346}}}
& \tc{The reference traverses image-reference sequences; generated SQL directly joins segmentation records and reproduces the accepted category counts.}
& \tc{Accepted aggregate outputs may be obtained without traversing reference-specific auxiliary fields.} \\[6pt]

\midrule
\multicolumn{4}{@{}l}{\textit{Gold SQL construction: implementation-path differences between released and constructed SQL}} \\
\midrule

\tc{\mbox{Gold SQL}}
& \tc{\mbox{\texttt{sf\_local003}}}
& \tc{Released SQL computes spending from payment records; constructed SQL aggregates item prices plus freight charges before customer-level RFM analysis.}
& \tc{Different monetary-value sources introduce different tables, fields, and aggregation paths.} \\
\noalign{\vskip 6pt}

\tc{\mbox{Gold SQL}}
& \tc{\mbox{\texttt{sf\_bq233}}}
& \tc{Released SQL reads R files directly from the contents table; constructed SQL uses a unified file--content join for both Python and R.}
& \tc{Language-specific versus unified retrieval paths use different sources for file-path fields.} \\
\noalign{\vskip 6pt}

\tc{\mbox{Gold SQL}}
& \tc{\mbox{\texttt{sf\_local075}}}
& \tc{Released SQL attributes purchases using event order; constructed SQL identifies purchase-containing visits through an event-dictionary join.}
& \tc{Sequence-based versus visit-based attribution changes both join and field dependencies.} \\

\bottomrule
\end{tabular}
\end{table*}

%% file: appendix/sensitive.tex
\section{Sensitivity to the Number of SQL Hypotheses}
\label{app:k_sensitivity}

\begin{figure}[t]
  \centering
  \includegraphics[width=\linewidth]{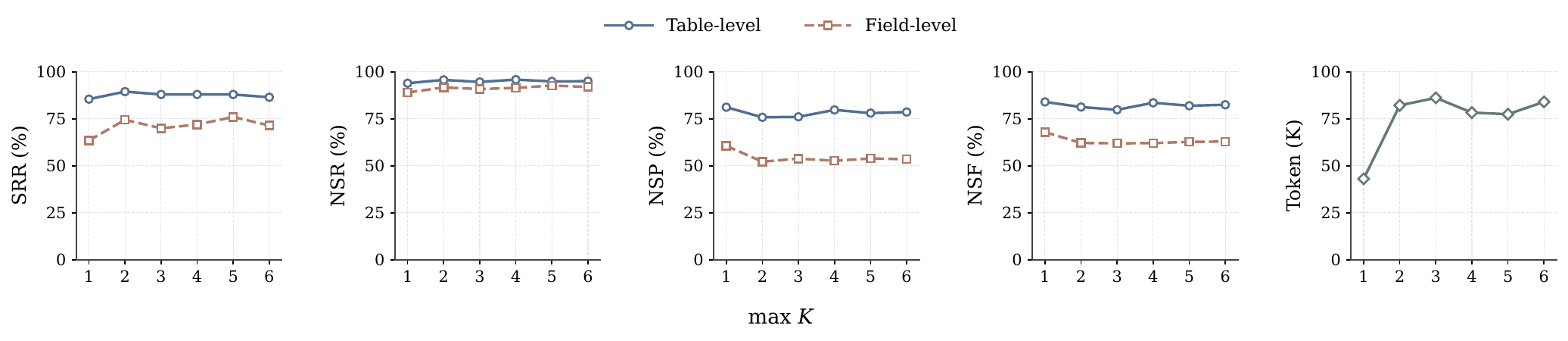}
  \caption{Sensitivity to the maximum number of SQL hypotheses $K$ on 200 examples randomly sampled from our Spider2-Snow evaluation set ($N=517$), using sampling seed 42. SRR, NSR, NSP, and NSF are reported at both table and field levels. Token denotes the average per-example token consumption during schema linking, measured in thousands. $K=1$ uses a single hypothesis, and $K=4$ is our default setting.}
  \label{fig:k_sensitivity}
\end{figure}

Multi-hypothesis grounding improves schema coverage but introduces a recall--precision trade-off.
Figure~\ref{fig:k_sensitivity} compares hypothesis budgets $K \in \{1,\ldots,6\}$ using Qwen3.8-27B on the same 200 examples randomly sampled from our 517-example Spider2-Snow evaluation set using seed 42. On this sample, all multi-hypothesis settings achieve higher SRR and NSR than $K=1$ at both table and field levels, supporting the benefit of considering alternative SQL realizations during grounding. However, these recall gains are accompanied by lower precision and F1, particularly at the field level. Most of the recall improvement is already obtained at $K=2$; further increases yield limited changes in NSR and non-monotonic changes in SRR.

Token consumption also varies non-monotonically with $K$, so a larger hypothesis budget does not necessarily incur a higher total cost. At $K=4$, table-level precision and F1 improve over $K=2$ and $K=3$, while token consumption is lower than at $K \in \{2,3,6\}$. On this sample, $K=5$ achieves the highest field-level SRR and NSR and the lowest token cost among the multi-hypothesis settings, while $K=4$ achieves higher table-level precision and F1 than $K=5$. We retain the default $K=4$, fixed before this analysis; these results illustrate the recall--precision--cost trade-off rather than establish a universally optimal hypothesis budget.

%% file: appendix/refinement.tex
\section{Analysis of Uncertainty-Guided Agentic Refinement}
\label{app:ugar_analysis}

We analyze how uncertainty-guided agentic refinement (UGAR) complements multi-hypothesis schema grounding through selective pruning and schema recovery on all 517 Spider2-Snow examples. Table~\ref{tab:ugar_diagnostics} reports the GLM-5.1 main experiment and the Qwen3.8-27B ablation separately. In the ablation, Full and \textit{w/o UGAR} share the same grounding output, isolating the effect of refinement on a fixed initial schema.

\input{tables/ugar}

\paragraph{Concentrated gains from selective refinement.}
UGAR's benefits are concentrated in a subset of examples, explaining why aggregate improvements appear modest. The selected tables and fields remain unchanged for 65.76\% of GLM-5.1 examples and 74.27\% of Qwen3.8-27B examples. Among examples whose schemas change, however, mean field F1 increases by approximately 4.35 and 3.28 percentage points, respectively. Full-dataset averages spread these improvements over many examples with unchanged schemas. Grouping examples by their observed refinement outcomes therefore reveals the gains associated with UGAR's schema adjustments more clearly.

\paragraph{Recovery beyond strict recall.}
Strong initial grounding leaves limited room for further gains in strict recall: GLM-5.1 already retains all reference fields on 486 of 517 examples before refinement. UGAR also provides useful recovery that is more visible in cases with incomplete initial schemas. On Qwen3.8-27B, among the 132 examples initially missing reference fields, UGAR raises mean field recall from 65.79\% to 68.20\%, improves recall on 15 examples, and restores complete recall on eight. Since SRR records only whether all reference fields are retained, partial recovery contributes to schema coverage without necessarily changing SRR. These results highlight the value of examining both aggregate completeness and the schema dependencies recovered on individual examples.

\paragraph{Illustrative pruning and recovery cases.}
Individual cases show how UGAR improves the schema context for specific query requirements. For \texttt{sf\_local163} in the GLM-5.1 run, UGAR removes five faculty attributes unrelated to the requested salary comparison, reducing the selection from 11 to six fields while retaining all four reference fields. This produces a more focused schema context without losing reference coverage. For \texttt{sf\_bq273} in the Qwen3.8-27B run, UGAR adds fields for sales price, inventory cost, and the inventory identifier needed for the profit calculation, raising field recall from 75\% to 100\%. Together, these cases illustrate UGAR's complementary roles in removing distracting attributes and recovering missing schema dependencies, showing how targeted refinement can provide useful improvements beyond those visible in aggregate scores.

%% file: tables/ugar.tex
\begin{table}[t]
\caption{Schema changes before and after uncertainty-guided agentic refinement (UGAR) on Spider2-Snow ($N=517$). Arrows indicate before $\rightarrow$ after refinement. SRR and NSF are field-level percentages; all counts refer to examples. Complete recall means retaining all reference fields. Refinement Token denotes average refinement tokens per example, in thousands.}
\label{tab:ugar_diagnostics}
\centering
\begin{tabular}{lcc}
\toprule
\textbf{Statistic} & \textbf{GLM-5.1} & \textbf{Qwen3.8-27B} \\
& Main experiment & Ablation experiment \\
\midrule
Field SRR & $94.00\rightarrow93.04$ & $74.47\rightarrow74.66$ \\
Field NSF & $35.16\rightarrow36.65$ & $62.47\rightarrow63.32$ \\
\midrule
Unchanged schema & 340 & 384 \\
Examples with fields removed & 175 & 95 \\
Examples with fields added & 4 & 39 \\
Field F1 improved / decreased & 175 / 2 & 104 / 27 \\
Complete recall before refinement & 486 & 385 \\
Complete recall gained / lost & 1 / 6 & 8 / 7 \\
UGAR Token & 16.27K & 11.93K \\
\bottomrule
\end{tabular}
\end{table}

%% file: appendix/tool.tex
\section{Tool Design}
\label{app:tool}

\begin{figure*}[t]
  \centering
  \includegraphics[width=\textwidth]{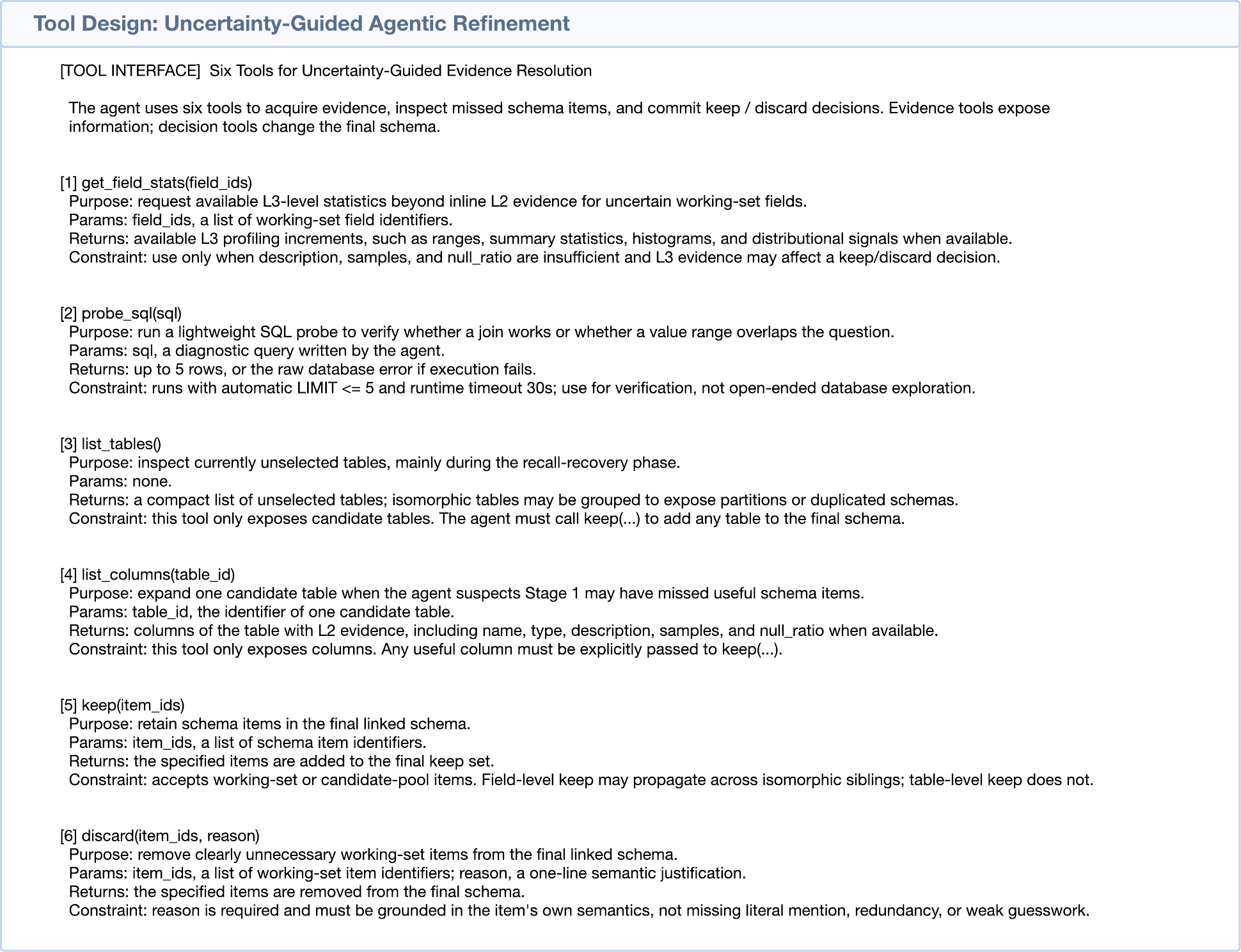}
  \caption{Tool design for uncertainty-guided agentic refinement. The six tools support targeted evidence acquisition, recall recovery, and keep/discard decisions under a recall-first policy.}
  \label{fig:tool}
\end{figure*}

Figure~\ref{fig:tool} summarizes the tool interface used in uncertainty-guided agentic refinement. The tools are designed to separate evidence acquisition from schema decisions. Specifically, \texttt{get\_field\_stats} and \texttt{probe\_sql} allow the agent to request stronger field-level evidence or lightweight SQL-level verification when the inline profile is insufficient. \texttt{list\_tables} and \texttt{list\_columns} support recall recovery by exposing schema items that were not selected in the initial grounding stage. Finally, \texttt{keep} and \texttt{discard} commit the refinement decisions. We make discard deliberately stricter than keep: a discard action requires an explicit semantic reason, while unresolved uncertain items are kept by default to preserve recall. The refinement agent provides a concrete implementation of our uncertainty-aware schema-linking formulation. Our focus is on using cross-path schema support to guide refinement and selective evidence acquisition. Future work can explore alternative agent architectures and tool configurations to further improve refinement quality and efficiency.

%% file: appendix/prompt.tex
\section{System Prompts}
\label{app:prompt}

\begin{figure*}[t]
  \centering
  \includegraphics[width=\textwidth]{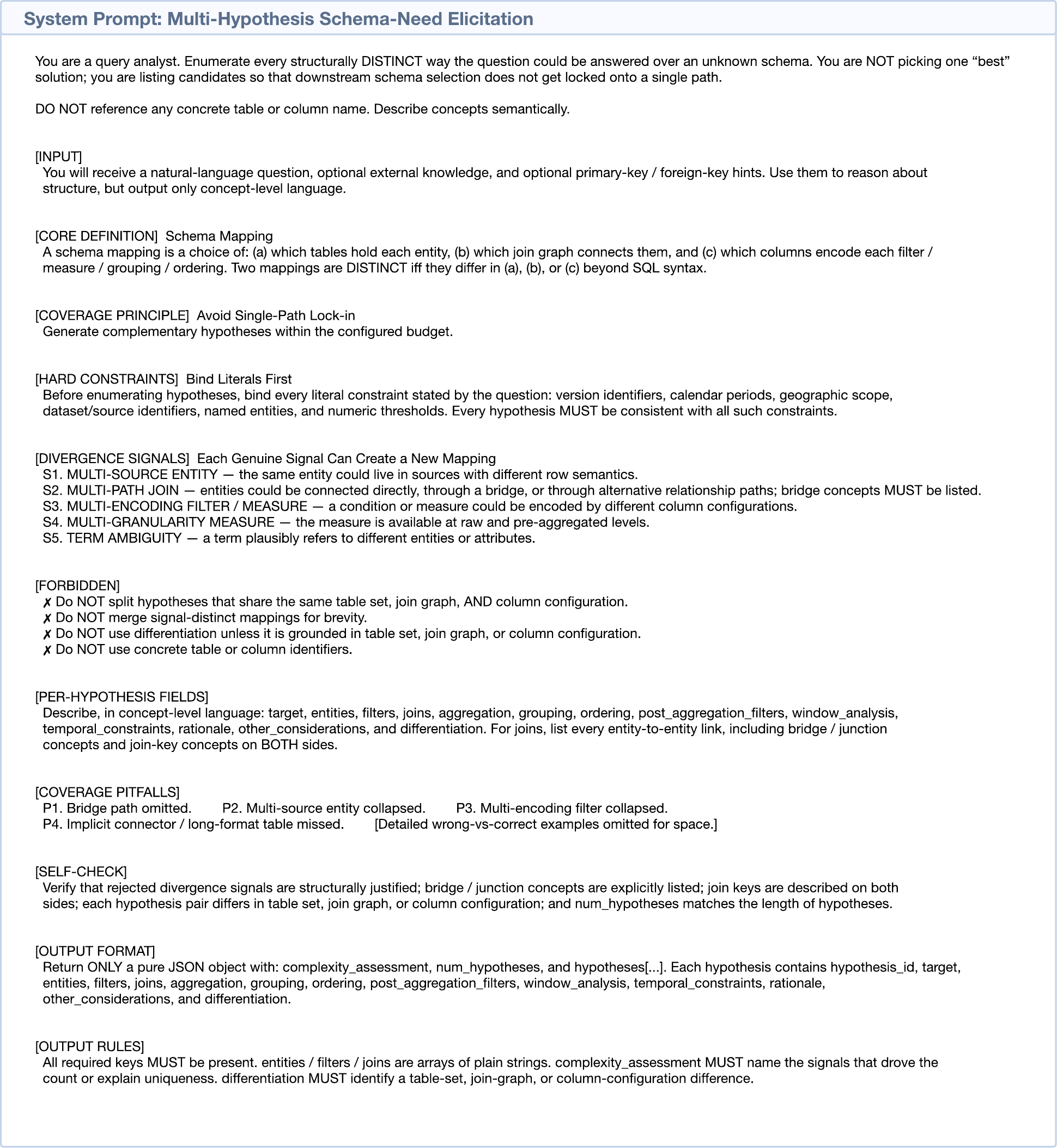}
  \caption{System prompt excerpt: multi-hypothesis schema-need elicitation.}
  \label{fig:prompt1}
\end{figure*}

\begin{figure*}[t]
  \centering
  \includegraphics[width=\textwidth]{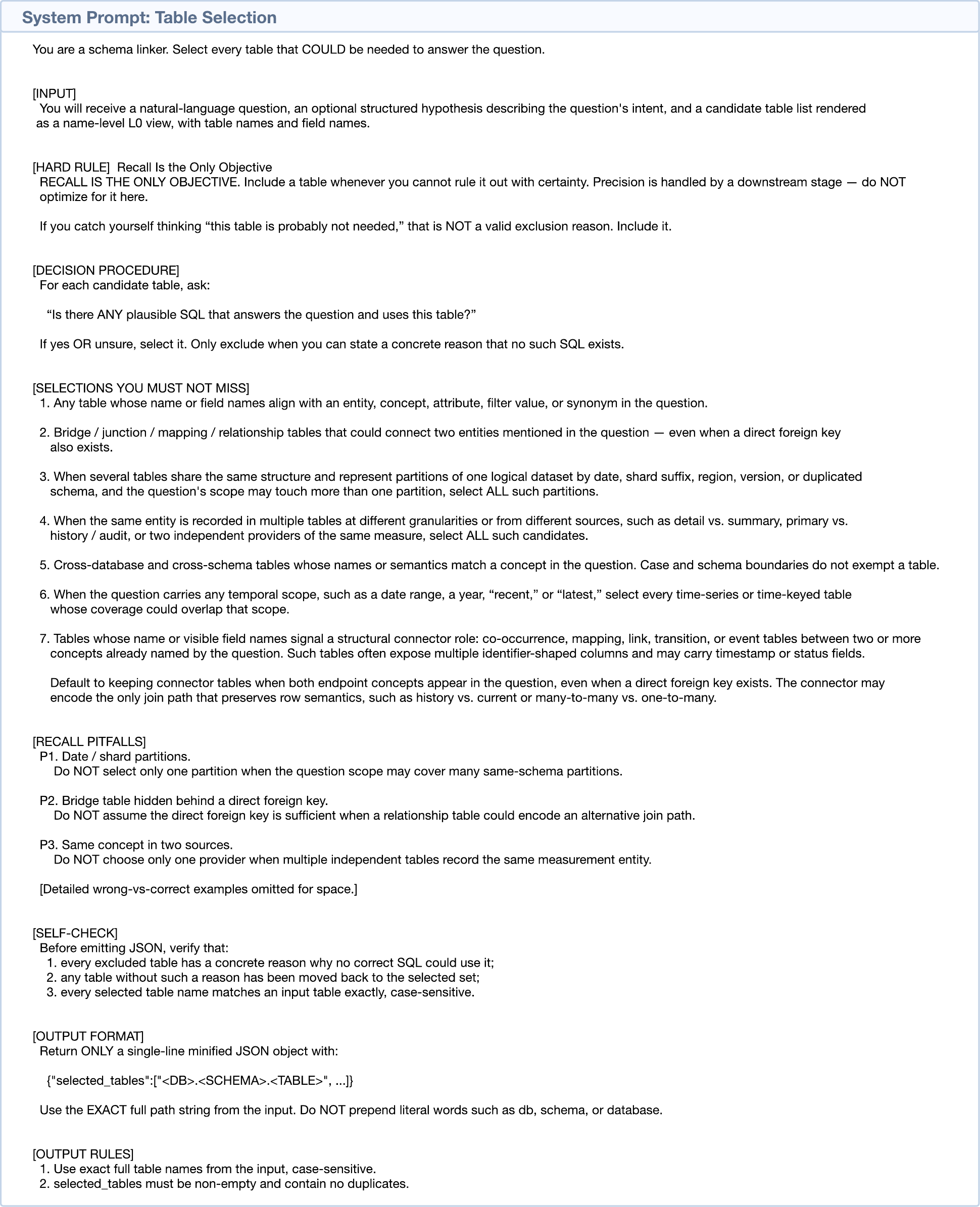}
  \caption{System prompt excerpt: table selection.}
  \label{fig:prompt2}
\end{figure*}

\begin{figure*}[t]
  \centering
  \includegraphics[width=\textwidth]{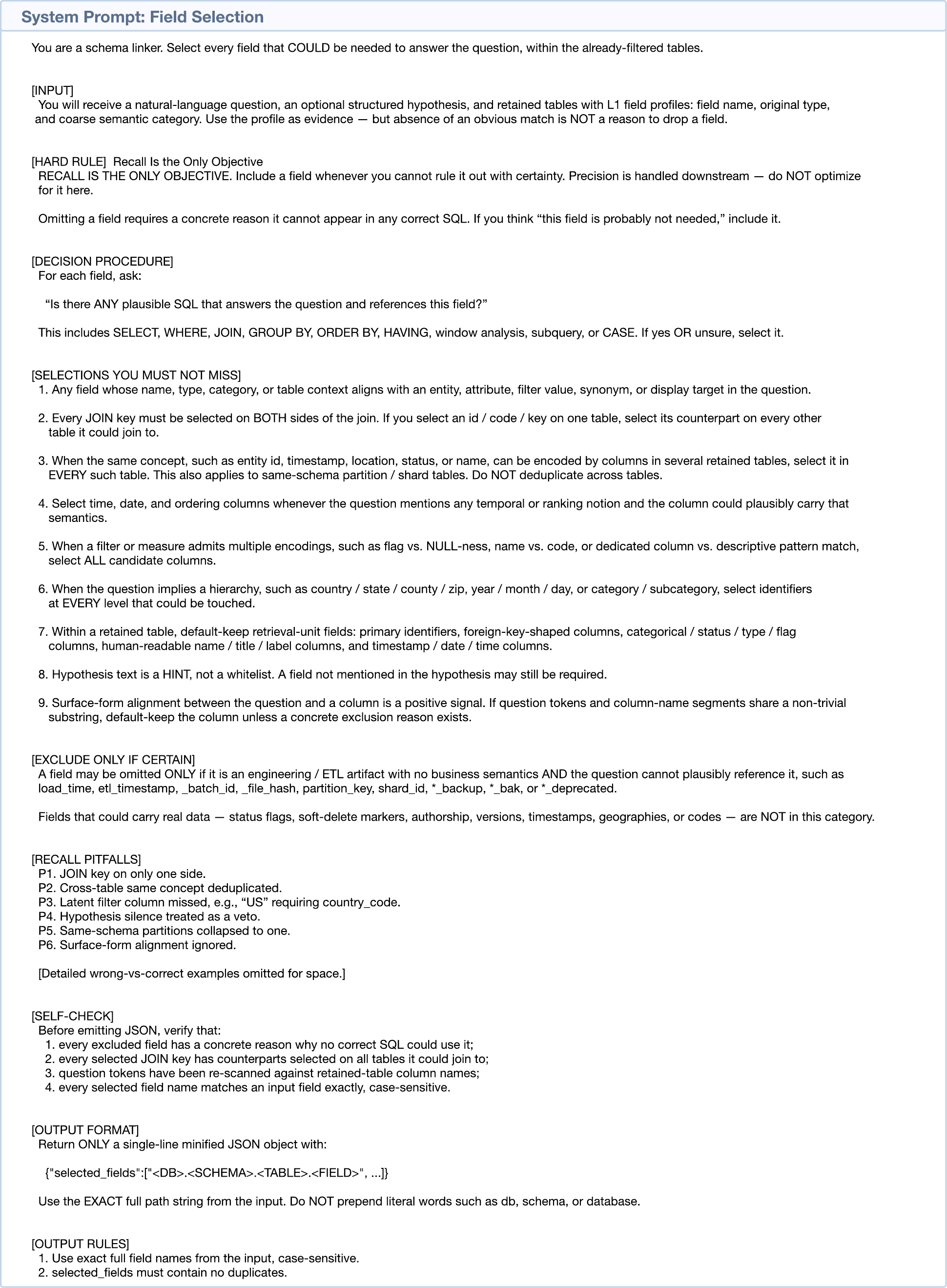}
  \caption{System prompt excerpt: field selection.}
  \label{fig:prompt3}
\end{figure*}

\begin{figure*}[t]
  \centering
  \includegraphics[width=\textwidth]{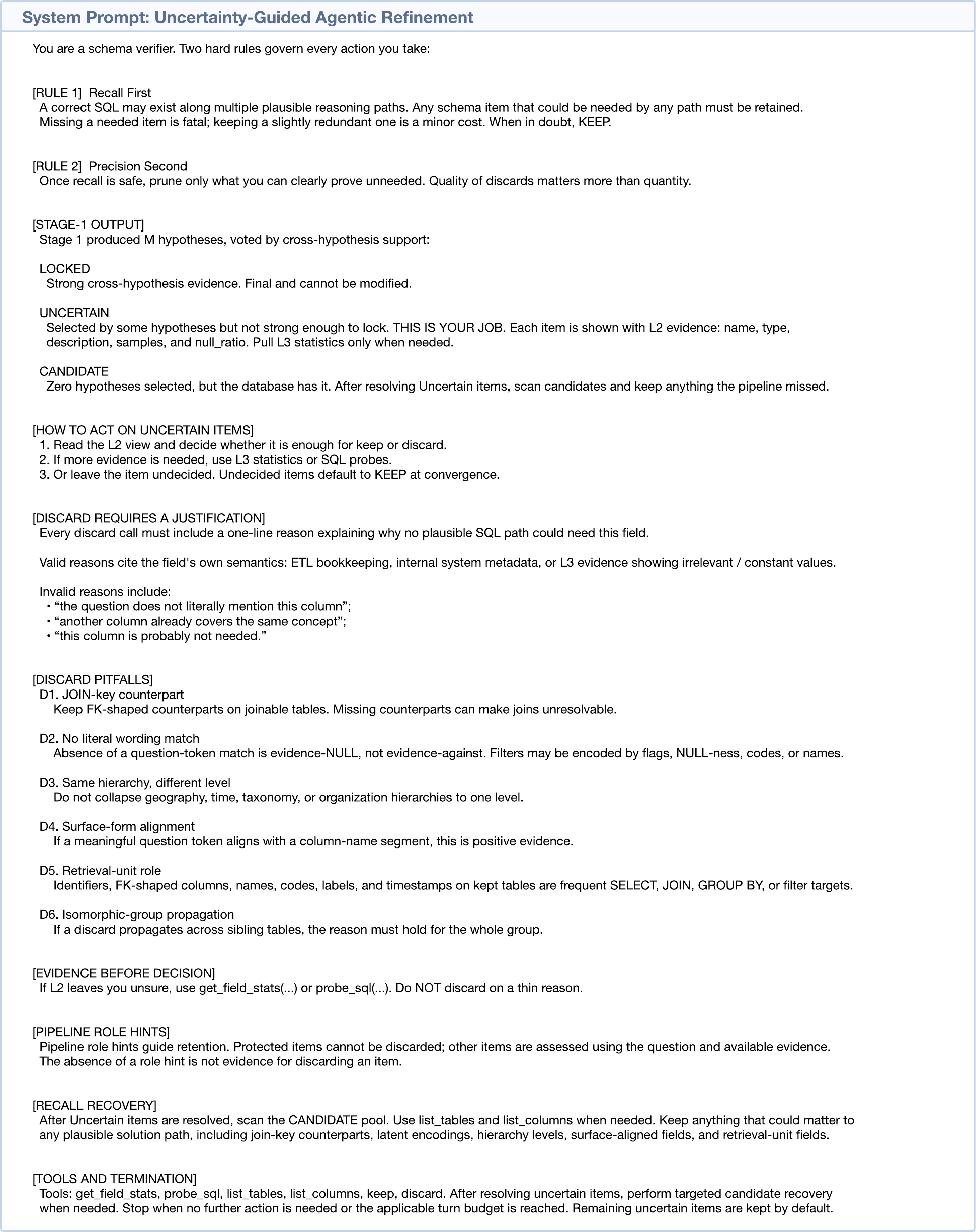}
  \caption{System prompt excerpt: uncertainty-guided agentic refinement.}
  \label{fig:prompt4}
\end{figure*}

Figures~\ref{fig:prompt1}--\ref{fig:prompt4} show the main system prompts used by EviLink: multi-hypothesis schema-need elicitation, table selection, field selection, and uncertainty-guided agentic refinement. For readability, we present faithful excerpts that preserve the operational instructions, constraints, and output requirements, while abbreviating long illustrative examples and verbose JSON templates.

\clearpage